# Where Language Models Fail Like Aphasic Patients — and Where They Don't

Yong Yang[1,*], Roger Newman-Norlund[4,3], Xiang Guan[1], Saeed Ahmadi[2], Regan Willis[1], Nadra Salman[5], Kalil Warren[5], Sophie Arheix-Parras[4], Srihari Nelakuditi[1], Rutvik H. Desai[4], Leonardo Bonilha[6], Christopher Rorden[4], Julius Fridriksson[2,3]

[1]*Department of Computer Science and Engineering, University of South Carolina;* [2]*Department of Communication Sciences and Disorders, University of South Carolina;* [3]*ALLT.AI, LLC;* [4]*Department of Psychology, University of South Carolina;* [5]*Linguistics Program, University of South Carolina;* [6]*Department of Neurology, School of Medicine, University of South Carolina*

[*]Corresponding author: yongy@email.sc.edu

## Abstract

The computational architecture of human word production—whether semantic-lexical retrieval and phonological encoding constitute dissociable processing stages—has been debated for decades using hand-designed models with small vocabularies and circular validation. We introduce computational architectural dissociation: using the known architectural properties of vision-language models (VLMs) to derive falsifiable predictions about which patient error types should and should not align with model errors under perturbation. Because VLMs perform picture naming through learned semantic competition over full vocabularies but lack phonological encoding entirely, perturbation-induced errors should match patient patterns for semantic-stage categories but not phonological ones. Testing LLaVA 1.6 (Vicuna-13B) against 88 patients with aphasia (PWA) across 175 Philadelphia Naming Test items (approximately 37 million model naming attempts, 50 perturbation seeds), we find that six of eight error categories show significant item-level alignment (Spearman $\rho$ = 0.32–0.50, all $p < 0.001$) while the two phonological categories—formal errors and abstruse neologisms—show no alignment ($\rho$ = 0.110 and 0.084, both n.s.), confirming the predicted dissociation. Output probability spectral analysis reveals that each error type occupies a distinct position in a six-dimensional spectral space, establishing a novel mechanistic interpretability tool. The dissociation is layer-specific (peaking at L22, 55% depth), severity-independent, and stable across seeds (split-half $r > 0.998$). Weight perturbation of four additional 7B VLMs spanning the tested architectural combinations produced correct-rate alignment of comparable magnitude ($\rho$ = 0.21–0.27). Whether the per-category dissociation itself depends on scale or on attention design requires a comparison within a single inference implementation, which is reported separately.

## 1. Introduction

### 1.1 Word production and the two-stage debate

Naming a picture—producing the word "cat" upon seeing one—is among the most frequent and seemingly simple language acts, yet it requires coordinating visual recognition, semantic retrieval, lexical selection, and articulatory planning (Indefrey & Levelt, 2004). When this process breaks down in aphasia, which affects approximately one million people in the United States (National Aphasia Association, 2019), the resulting errors provide a window into its internal organization (Dell et al., 1997). Semantic errors (saying "dog" for "cat") suggest disrupted retrieval from a semantic neighborhood; formal errors (saying "mat" for "cat") implicate phonological selection; mixed errors combine both relationships; and no-response errors suggest complete retrieval failure.

The dominant theoretical framework is Dell's interactive activation model (Dell et al., 1997; Foygel & Dell, 2000), which posits two processing stages governed by connection-strength parameters: a semantic-lexical stage (s) governing selection from semantic competitors, and a phonological stage (p) governing phoneme assembly. This two-stage architecture has been supported by computational modeling with small interactive networks (Dell et al., 1997; Martin et al., 1994), by analyses of error type distributions across patients with varying severity, and by lesion-symptom mapping studies (Fridriksson et al., 2018) showing partial dissociations between lesion sites associated with semantic versus phonological errors.

Aphasia is the most common acquired language disorder, arising in roughly one-third of acute stroke cases (Engelter et al., 2006), and naming impairment is among the symptoms patients and families find most disabling because it disrupts everyday conversation and social participation. Within Dell's framework, impairment at the first (semantic-lexical) stage produces semantic errors, mixed errors, and unrelated real-word substitutions, whereas impairment at the second (phonological) stage produces formal paraphasias, phonologically related nonwords, and abstruse neologisms; no-response errors can arise from severe impairment at either stage. Competing accounts—Indefrey and Levelt's strictly serial model, Dell's interactive model, and Hickok and Poeppel's (2007) dual-stream framework—make different predictions about the nature and locus of the stage boundary, and patient error patterns alone have not adjudicated among them.

A further limitation is that Dell's model parameterizes phonological impairment with a single weight parameter (p), predicting that all phonological error types—formal paraphasias, nonword errors, and neologisms—should be affected uniformly by phonological-level damage. This prediction has never been tested at the item level, because classical models lack the scale to compare item-specific vulnerability across phonological subtypes. Whether phonological errors form a unitary category or comprise functionally distinct subtypes with different vulnerability mechanisms therefore remains an open empirical question that the present work is positioned to address.

However, the framework faces a circularity problem: the same error patterns used to define the two stages also serve as the primary evidence for their separability. Classical computational models share three fundamental limitations. First, they are explicitly designed with separate semantic and phonological stages, making their confirmation of stage separability non-independent. Second, they operate over vocabularies of 30–40 words, far below the scale needed for item-level predictions across a 175-item clinical test. Third, they require researcher-specified semantic and phonological similarity matrices, introducing assumptions about the very representations whose organization is being tested. These limitations constrain what the models can reveal about the architecture they were built to embody (Supplementary Note 5). Independent evidence from a system not designed with two stages is needed to break this circularity.

Our work stands in the tradition of computational neuropsychology pioneered by Plaut and Shallice (1993), who used lesioned connectionist networks to model deep dyslexia. Like Dell's naming models, those networks were purpose-built to instantiate specific theoretical commitments. Our approach differs in a decisive respect: LLaVA was not designed to model word production. It was trained for general vision-language tasks, and its possession of semantic competition together with its complete absence of phonological encoding are incidental features of its design rather than deliberate theoretical choices. This independence from the hypothesis under test is precisely what eliminates the circularity inherent in purpose-built models. The brain–AI alignment literature (Caucheteux & King, 2022; Goldstein et al., 2022; Schrimpf et al., 2021) provides a methodological precedent but addresses a different question—representational similarity during comprehension in healthy participants—and cannot test whether the shared representations reflect modular architecture.

## 1.2 Vision-language models as architectural probes

Modern vision-language models (VLMs) resolve all three limitations simultaneously. Models such as LLaVA (Liu et al., 2024) combine a pre-trained vision encoder (CLIP ViT-L/14) with a large language model (Vicuna-13B) to perform picture naming through learned token prediction over full vocabularies (~32,000 BPE tokens), without explicit stage decomposition and without researcher-specified similarity structure. The semantic competition mechanism emerges naturally from training: when naming an image, the output layer produces a probability distribution over all tokens, with semantically related words receiving systematically higher probabilities than unrelated ones. Perturbing internal representations disrupts this competition, producing naming errors whose types depend on which processing levels are affected.

The correspondence between attention-based selection and Dell's semantic-lexical competition rests on three structural parallels rather than superficial output resemblance. First, both systems implement competitive selection among candidates: semantic features activate multiple lexical nodes that compete on activation level in Dell's model, while the attention mechanism weights candidate tokens by contextual similarity and the softmax amplifies high-scoring candidates. Second, both exhibit graded interference: semantic errors arise when a competitor's activation exceeds the target's in patients, and when a competitor token's attention-weighted probability exceeds the target's in the model. Third, both show item-specific vulnerability: items with dense semantic neighborhoods are more error-prone in patients (Mirman, 2011) and in the model, because competition is fiercer.

Critically, VLMs lack a phonological encoding stage entirely. Output is generated through byte-pair encoding (BPE) tokenization (Sennrich et al., 2016) that segments text based on corpus frequency statistics, not phonological similarity. When a VLM produces an error, the output token is selected through disrupted semantic-lexical competition, not phonological neighborhood interference. Formal analysis confirms that BPE similarity and phonological similarity are essentially uncorrelated for PNT targets (Supplementary Note 5).

A natural objection is that BPE tokenization, though not phonological by design, might capture phonological structure indirectly, since English orthography correlates with phonology at roughly 80% regularity (Venezky, 1970). We acknowledge this partial overlap but argue it is insufficient to constitute phonological encoding. Phonological encoding requires three operations BPE cannot perform: activating individual phonemes from a stored phonological representation (BPE operates on character n-grams, not phonemes); assembling phonemes into a syllabic frame with stress and prosody (BPE has no syllable or stress representation); and accessing phonological neighbors—words differing by a single phoneme—as competitors during encoding (BPE has only character-overlap statistics). The critical test is whether item-specific phonological vulnerability is shared between patients and the model; our finding that formal errors do not align (Section 3.4) confirms that incidental character overlap is not functionally equivalent to phonological encoding.

This architectural property enables what we term computational architectural dissociation, adapting the logic of classical neuropsychological dissociation to artificial systems: (i) identify a known architectural absence in a computational model—here, phonological encoding; (ii) derive predictions about which patient error types should and should not align with model errors; (iii) test these predictions against a large clinical dataset. If the alignment boundary falls precisely where the architectural boundary predicts—between semantic-stage and phonological-stage error types—this constitutes independent, architecturally grounded evidence that the two stages are functionally dissociable. Unlike brain–AI alignment studies examining representational similarity (Schrimpf et al., 2021; Goldstein et al., 2022), our approach uses what a model architecturally lacks to make positive predictions about cognitive architecture.

### 1.3 Predictions and experimental design

Before examining any patient data, we derive predictions from LLaVA's architecture. Group A (predicted alignment): Correct responses, semantic errors, mixed errors, and unrelated errors are all generated through semantic-lexical competition, which LLaVA possesses; their item-level rates should correlate with patient rates. Group B (predicted non-alignment): Formal errors require access to phonological neighbors; abstruse neologisms are word-like but phonologically unrelated to targets. Both depend on phonological processing that LLaVA lacks. Group C (ambiguous prediction): Nonword errors are phonologically defined but may be generated at the semantic-lexical level; no-response errors could arise from either stage.

These predictions are falsifiable in both directions. If formal errors show strong item-level alignment despite LLaVA's lack of phonological encoding, the architectural dissociation argument fails. We test predictions with three convergent evidence levels: (1) spectral characterization of the error taxonomy using six output probability distribution observables, (2) item-level alignment across 175 PNT items, and (3) layer-specific localization within the 40-layer architecture.

If, instead, all categories failed to align, the perturbation framework would lack external validity; and if the alignment boundary fell between frequent and rare error types rather than between semantic-lexical and phonological categories, the architectural explanation would be refuted. The prediction is thus directional and pre-registered in logic: six categories should align and two should not, and the boundary should track processing requirements rather than statistical properties of the categories.

The data comprise 88 patients with chronic post-stroke aphasia from the POLAR cohort (Fridriksson et al., 2018), tested on 175 PNT items across 1,007 sessions (165,375 scored records), and LLaVA 1.6 tested with 50 perturbation seeds across the full perturbation grid (approximately 37 million naming attempts). To contextualize: Schwartz et al.'s (2006) landmark Dell model validation used 94 participants with approximately 16,450 scored responses; our dataset provides approximately 10 times more patient data and approximately 2,200 times more model evaluations. The output spectral analysis—six observables characterizing how probability distributions degrade—constitutes a methodological contribution applicable to any setting involving perturbed language model performance.

This work follows a line of studies that established the phenomenon it now seeks to explain. Perturbing a multimodal language model reproduces the naming-error profiles of individual people with aphasia (Yang et al., 2026a); the perturbation parameters can be recovered from a profile, making the mapping invertible (Yang et al., 2026b); lesion-symptom mapping serves as an external reference for what a perturbation has done (Fridriksson et al., 2026); and subtraction analysis localizes error classes along the layer axis (Guan et al., 2026). Three things distinguish the present study. It asks about items rather than proportions: the earlier work matched how often each error type occurs, whereas we ask whether the same pictures are difficult, and a model can reproduce a patient's error proportions exactly while failing on an unrelated set of items. It restores a distinction the earlier work collapsed: the classifier is trained on eight categories and its eight-category output is where both studies begin, but the seven-category scheme used previously merges nonwords into neologisms, and those two categories are the object of this study. Eight-category coding is available for the 88-patient POLAR cohort rather than the 278 of the earlier work, so the trade is resolution against sample size, and the distinction earns its cost: nonwords and neologisms sit a normalized spectral distance of 1.396 apart and behave differently in their alignment with patients, neither of which survives the merge. And it states a prediction before testing it, deriving from the model's architecture which categories should align and which should not; what the earlier work reported as an anomaly, formal paraphasia failing to emerge at clinically comparable proportions, is here a prediction, joined by neologism once the collapsed category is separated.

## 2. Methods

### 2.1 Patient data and scoring

Participants were 88 persons with chronic post-stroke aphasia (>6 months) from the POLAR longitudinal cohort (Fridriksson et al., 2018), spanning the full severity range (WAB-AQ 11.2–93.4, mean 60.0). Participants completed the 175-item Philadelphia Naming Test (PNT; Roach et al., 1996) across 1,007 sessions (median 12 per patient), yielding 165,375 scored item records classified into eight categories: correct, semantic, formal, mixed, unrelated, nonword, neologism, and no-response. Detailed inclusion criteria, demographic data, and the session-weighting scheme are in SI Methods and Supplementary Note 1.

Participants were enrolled at the University of South Carolina Center for the Study of Aphasia Recovery (C-STAR; NIH/NIDCD P50 DC014664) and spanned the major WAB-classified subtypes (Broca's, Wernicke's, conduction, anomic, global, transcortical), with a small subset classified as non-aphasic on the WAB, ensuring the findings are not specific to any single aphasia profile. The longitudinal design—patients tested at multiple time points across two treatment arms (Phonological Components Analysis and Verb Network Strengthening Treatment)—provides both cross-sectional power and the opportunity to examine recovery trajectories. Extended inclusion criteria, demographics, subtype composition, and the comparison of session-weighted versus unweighted item rates are reported in SI Methods.

### 2.2 VLM perturbation

We used LLaVA 1.6 (Liu et al., 2024) with the Vicuna-13B backbone (Zheng et al., 2023), a 40-layer transformer with dense multi-head attention (40 query/key-value heads) and sequential attention-MLP layout. Unit-level weight perturbation selected X% of hidden units and applied multiplicative Gaussian noise ($w' = w(1 + N(0, \sigma))$) to all aligned weight matrices. The grid spanned 40 layers × 10 modification levels (10–100%) × 10 noise levels (1.1–2.0) × 185 items × 50 seeds, approximately 37 million naming attempts in total. Perturbation rationale and parameters are in Supplementary Note 3.

The parameter ranges were chosen to span the full clinical spectrum: the mildest conditions reduce overall correct rate by roughly 5–10 percentage points from baseline, while the most severe drive it to 10–25% with predominantly no-response and neologism outputs. Gaussian noise was preferred over dropout, pruning, or activation zeroing because it preserves the statistical structure of activations while reducing signal-to-noise ratio (a plausible analog of Dell's uniform connection-strength reduction), is graded and continuous (enabling fine-grained dose–response exploration), and can be applied layer-by-layer (enabling the layer-stratified analysis central to the dissociation test). A critical implementation detail is that the output probability distribution must be captured immediately after the first forward pass, before the autoregressive loop overwrites the activations; delayed extraction silently corrupts the spectral observables. Full rationale appears in SI Methods and Supplementary Note 3.

### 2.3 Classification, spectral analysis, and alignment

Model responses were classified using a fine-tuned DeBERTa classifier (He et al., 2021; F1 = 0.78–0.99 per category; SI Table S3). Six spectral observables were computed from each output probability distribution: spectral flatness (SF), Zipf exponent (α), participation ratio (PR), KL divergence ($D_{KL}$), Shannon entropy (H), and spectral contrast ratio (SCR). These observables characterize distribution shape and are defined formally in SI Methods. The error classifier and the spectral analysis are formally distinct—the classifier labels responses, while the spectral observables characterize the output distribution that produced them; their relationship is derived formally in SI Note 6.

The six observables were chosen to capture distinct aspects of distribution shape rather than a single axis of uniformity. Spectral flatness and Shannon entropy both index overall uniformity but emphasize different regions (SF is dominated by the geometric mean and thus the tail; H by the bulk). The Zipf exponent captures power-law tail steepness, distinguishing a few-dominant-token distribution from a many-competitor one. Participation ratio gives a directly interpretable count of effective competing tokens. KL divergence from uniform summarizes overall confidence in a single number, and SCR captures the winner-take-all character of selection. All observables are computed over the full ~32,000-token vocabulary rather than a truncated set, because the tail structure carried by PR, α, and SF would otherwise be lost. Complete formal definitions appear in SI Methods.

We adopt the eight-category scheme over the common seven-category alternative, which merges phonologically related nonwords into the neologism class, on spectral grounds: nonword outputs (spectral flatness ≈ 0.75) and neologisms (≈ 0.83) are separated by a normalized spectral distance of roughly 1.4 and therefore occupy clearly distinct regions of spectral space. Collapsing them would discard exactly the nonword–neologism contrast on which the phonological-stage dissociation of Section 4.2 depends.

Item-level alignment was quantified by Spearman ρ between model and patient per-item error rate vectors (175 items), with significance assessed via permutation tests (10,000 permutations). Patient rates were session-weighted averages; model rates were seed-averaged proportions. Multiple comparison correction was not applied because the eight tests address a priori predictions with pre-specified directionality. The 50-seed design enables split-half reliability analysis (odd-seed vs. even-seed item vectors).

Two features of the statistical design deserve emphasis. The aligned categories' Spearman ρ of 0.32–0.50 correspond to roughly 10–25% shared item-level variance—substantial by the standards of individual-differences research given the gulf between a 13-billion-parameter transformer and a human brain recovering from stroke. And the 50-seed design permits treating each seed as an independent replication, so that psychometric reliability statistics standard in clinical assessment (split-half with Spearman–Brown correction) can be applied to a computational model; coefficients exceeding 0.998 confirm that the alignment is a stable, reproducible property of the architecture rather than an artifact of any single noise realization.

## 3. Results

### 3.1 Perturbation produces a spectrally principled error taxonomy

Perturbation of LLaVA's internal representations produced naming errors spanning all eight PNT categories, with error types depending systematically on perturbation intensity and layer. Analysis of output probability distributions revealed that each error type occupies a distinct region in six-dimensional spectral space (Fig. 1). The spectral hierarchy, ordered by spectral flatness, runs: correct (SF = 0.271, highly peaked) through semantic (0.385), mixed (0.388), unrelated (0.592), formal (0.648), no-response (0.700), nonword (0.750), to neologism (0.831, nearly uniform).

This ordering is continuous: the eight categories occupy distinct, separable regions of a smooth manifold (Fig. 2) whose geometry is determined by the perturbation's effect on the model's internal probability distributions. The spectral hierarchy maps the traditional linguistic taxonomy onto a computational axis: correct responses correspond to sharply peaked distributions (high confidence), semantic errors to moderately peaked distributions (competition among related items), and degraded categories to progressively flatter distributions. This mapping provides a mechanistic bridge between observable behavior and internal processing state, constituting a novel tool for mechanistic interpretability applicable beyond the present study.

The relationship between perturbation intensity and spectral observables follows smooth, monotonic dose-response curves (SI Figures S1–S3), ruling out catastrophic or discontinuous effects and confirming that perturbation creates a continuous landscape of impairment levels.

Beyond validating the taxonomy, the six spectral observables constitute a new class of interpretable objects for mechanistic analysis. Whereas standard mechanistic interpretability characterizes internal structures—attention heads, circuits, or individual neurons—spectral analysis characterizes the output distribution, the final common pathway through which all internal processing is expressed, and does so without access to internal activations. Each observable summarizes the model's decision state at the moment of word production: the degree of lexical competition, the steepness of the candidate hierarchy, the count of effective competitors, and the overall confidence of selection. The spectral signature of an error type is therefore not a classifier label but a quantitative characterization of the distribution from which the erroneous output was sampled—an interpretable bridge between the opaque dynamics of a 13-billion-parameter transformer and the clinically observable categories of naming failure.

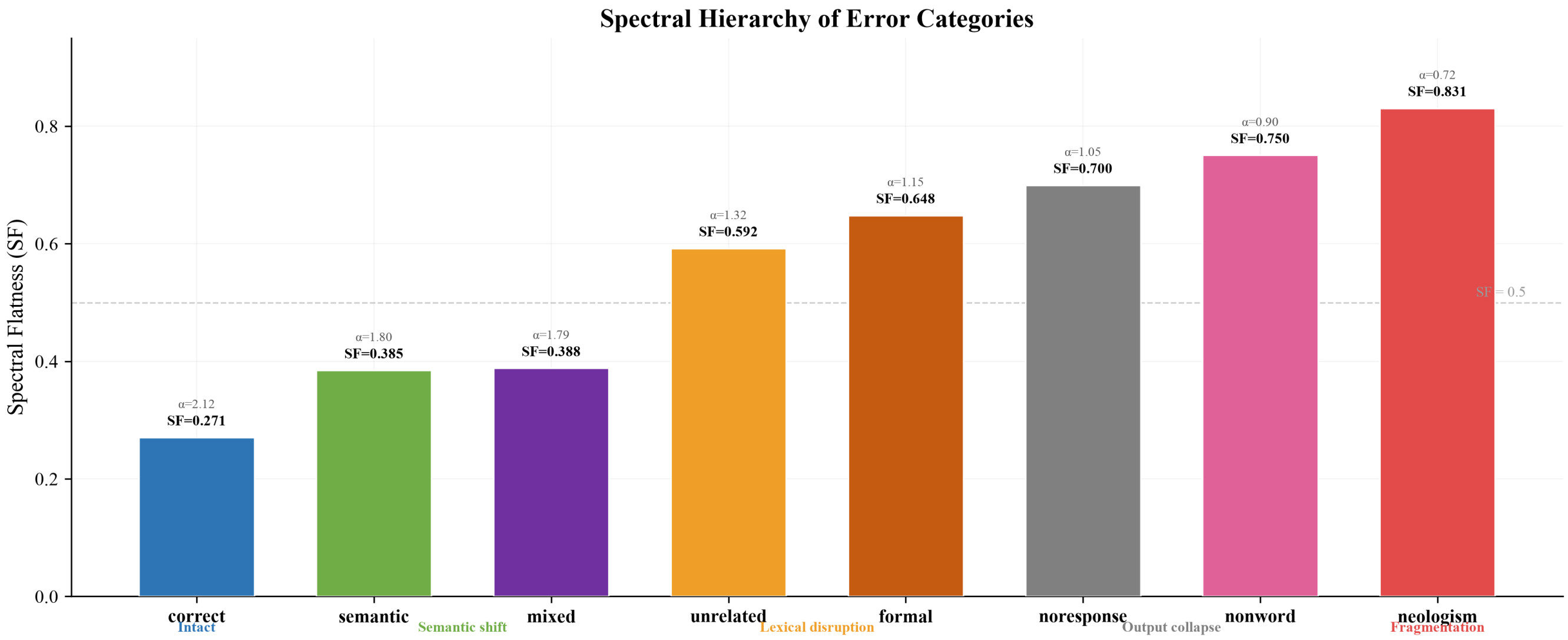

***Figure 1.*** *Spectral hierarchy of naming error types. Each category occupies a distinct spectral position, from peaked (correct) to uniform (neologism/no-response). Error bars: 95% CI across 50 seeds.*

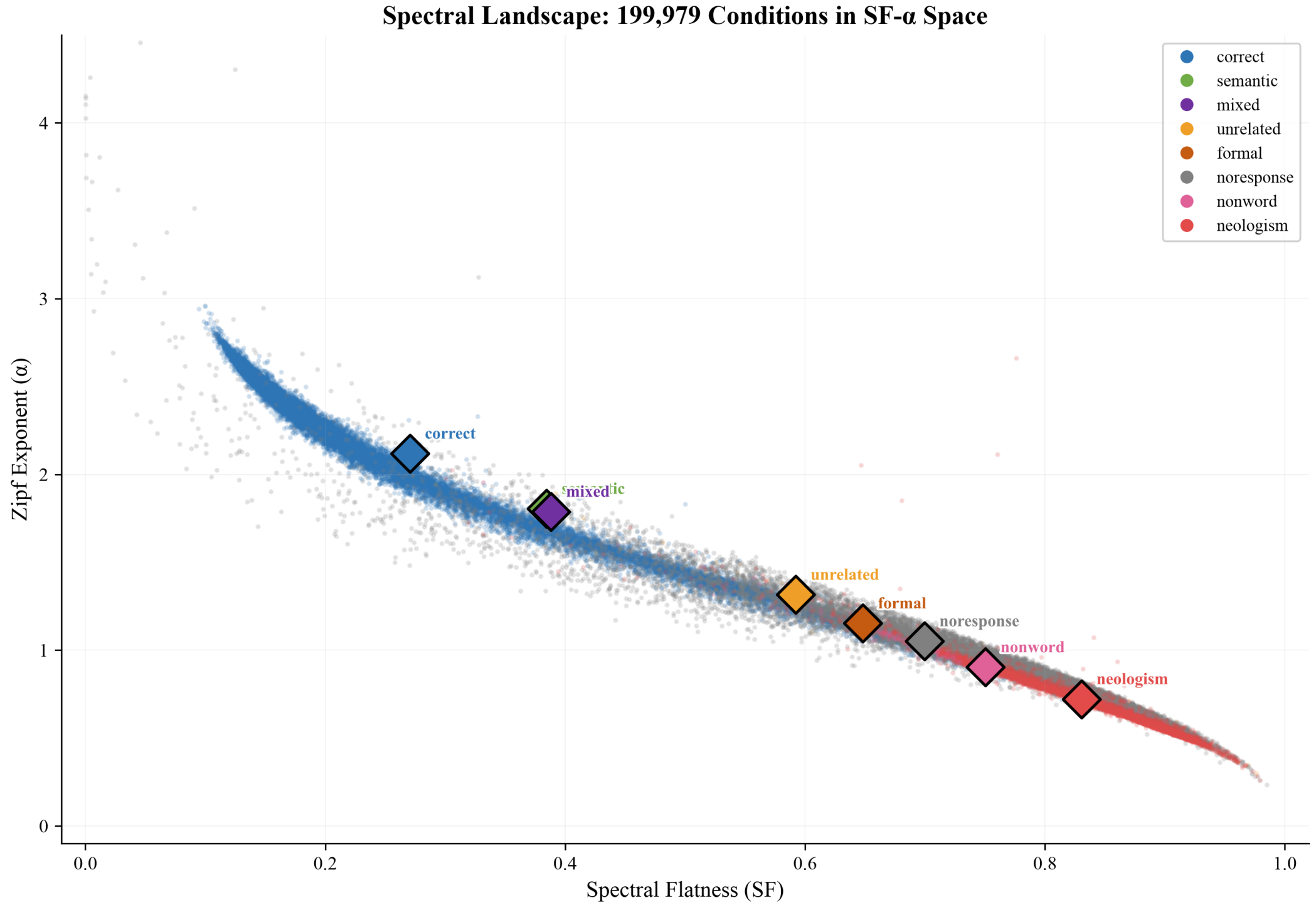

***Figure 2.*** *Spectral landscape: 200,000 conditions in SF-α space, colored by dominant error category. Category centroids (diamond markers) trace a smooth curve from correct (high α, low SF) through intermediate categories to neologism (low α, high SF). The dense blue cloud at SF < 0.3 reflects the predominance of correct responses under mild perturbation. Each point represents one perturbation condition × seed combination; the*

*continuous manifold confirms that error types emerge from smooth degradation of the output distribution rather than discrete transitions. N = 199,979 condition-seed pairs from 50 seeds.*

### 3.2 Spectral distance reveals error structure

Pairwise spectral distances were computed using z-score normalized Euclidean distance in six-dimensional space (Fig. 3). Z-score normalization is essential: the spectral contrast ratio has a range approximately ten times larger than other observables and would dominate without standardization. Three spectral clusters emerged: correct responses (isolated), semantic-lexical errors (semantic and mixed, closely spaced), and degraded outputs (nonword, unrelated, neologism, no-response).

The spectral distance between error pairs predicted their co-occurrence probability with remarkable strength: proportion-weighted Spearman r = −0.921 ($p < 0.0001$, 85% variance explained; Fig. 4). Spectrally close error types frequently co-occur at the same perturbation conditions. A key finding: mixed errors are spectrally much closer to semantic errors (normalized distance 0.281) than to formal errors (2.80), consistent with generation through semantic competition with incidental phonological overlap. The strength of this distance–co-occurrence relationship (Fig. 4) establishes the spectral observables as a meaningful coordinate system for the error taxonomy.

The z-score normalization applied before distance computation is essential and interpretable: without it, SCR—whose range spans roughly three orders of magnitude—dominates the Euclidean metric, making all distances effectively proportional to SCR differences. After independent standardization of each observable, categories separated by normalized distance greater than 2.0 are clearly distinguishable in spectral space while those below 0.5 spectrally overlap. The mixed–semantic distance of 0.281 falls well below 0.5 (substantial overlap), whereas the mixed–formal distance exceeds 2.0 (clear separation), a roughly 9:1 asymmetry. The predictive power of spectral distance over co-occurrence (r = −0.921) shows that spectral space is not merely a convenient visualization but a functionally meaningful representation: two conditions with similar signatures produce similar errors because they correspond to similar internal states, even when their perturbation parameters differ.

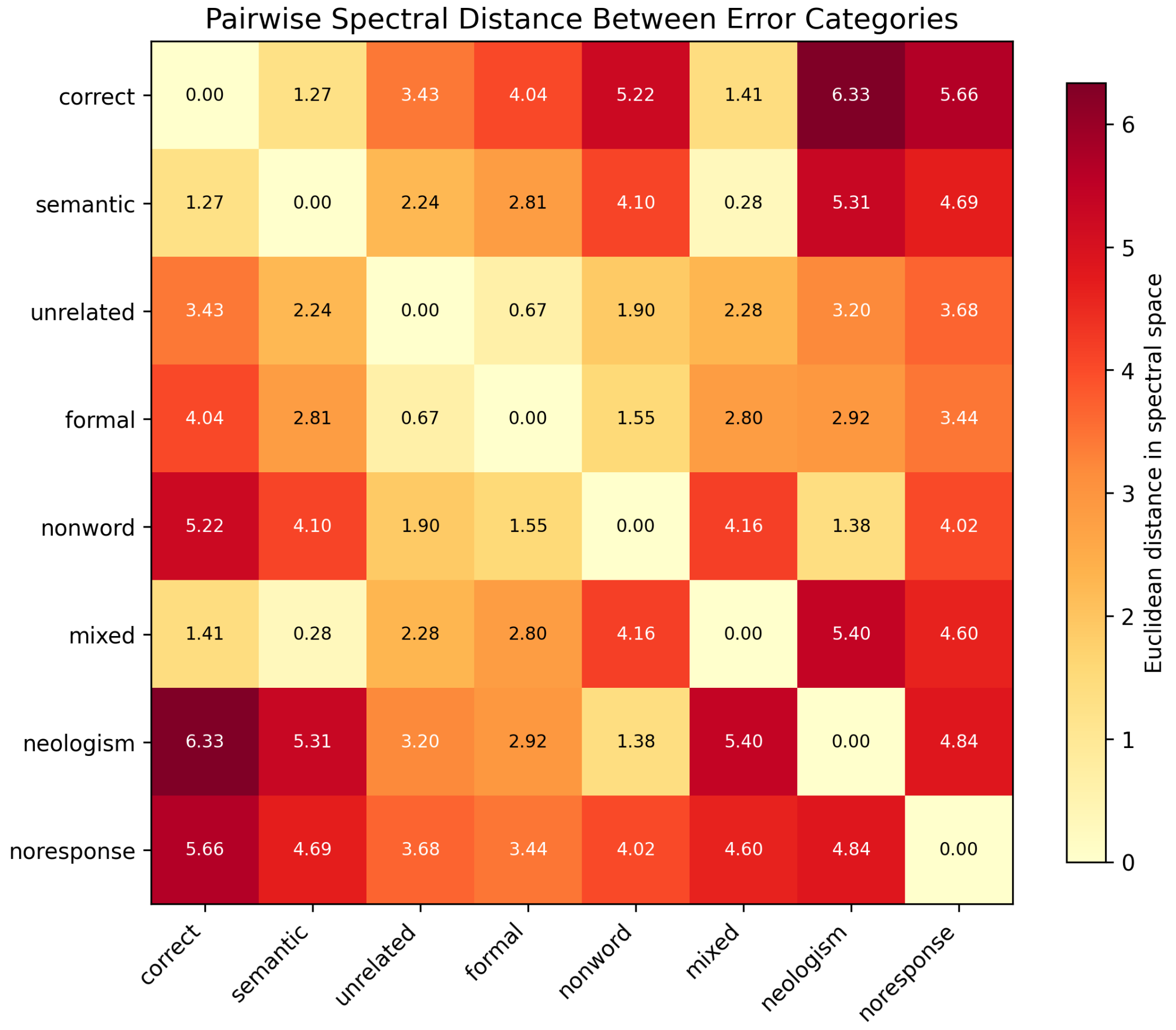

***Figure 3.*** *Spectral distance matrix between error categories (z-scored Euclidean). Three clusters emerge. Mixed-to-semantic distance (0.281) is much smaller than mixed-to-formal (2.80), indicating mixed errors are semantic variants in VLMs.*

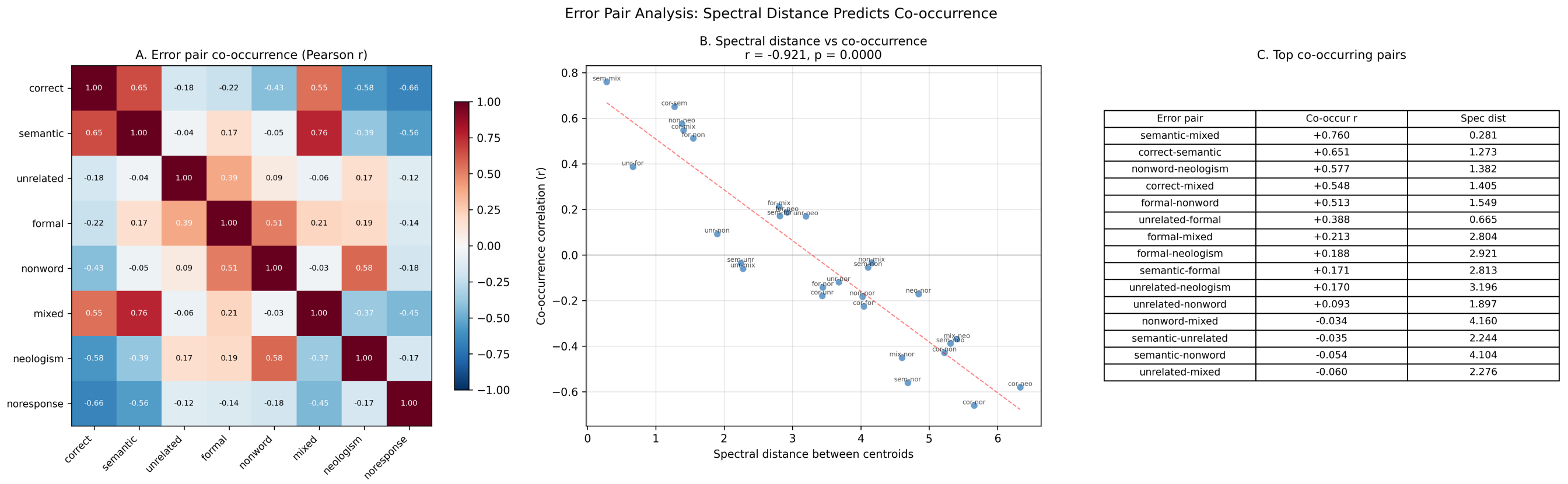

| Error pair | Co-occur r | Spec dist |
|---|---|---|
| semantic-mixed | +0.760 | 0.281 |
| correct-semantic | +0.651 | 1.273 |
| nonword-neologism | +0.577 | 1.382 |
| correct-mixed | +0.548 | 1.405 |
| formal-nonword | +0.513 | 1.549 |
| unrelated-formal | +0.388 | 0.665 |
| formal-mixed | +0.213 | 2.804 |
| formal-neologism | +0.188 | 2.921 |
| semantic-formal | +0.171 | 2.813 |
| unrelated-neologism | +0.170 | 3.196 |
| unrelated-nonword | +0.093 | 1.897 |
| nonword-mixed | -0.034 | 4.160 |
| semantic-unrelated | -0.035 | 2.244 |
| semantic-nonword | -0.054 | 4.104 |
| unrelated-mixed | -0.060 | 2.276 |

***Figure 4.*** *Error pair analysis: spectral distance predicts co-occurrence. (A) Pairwise co-occurrence matrix (Pearson r) across 200,000 perturbation conditions. Semantic-mixed (r = 0.76) and correct-semantic (r = 0.63) are the most co-occurring pairs. (B) Spectral distance between category centroids vs empirical co-occurrence (r = −0.921, p < 0.0001). Error categories with similar spectral signatures co-occur in the same perturbation conditions; spectrally distant categories are anti-correlated. (C) Top co-occurring pairs ranked by co-occurrence strength with corresponding spectral distances.*

## 3.3 Six of eight error categories show significant item-level alignment

Having established that perturbation produces a principled error taxonomy, we test whether LLaVA's item-specific error patterns correspond to those of real patients. For each error category, we computed Spearman ρ between the model's per-item error rate vector (175 items, averaged across 50 seeds) and the POLAR patients' session-weighted per-item rate vector (Table 1).

The strongest alignment was for correct rate ($\rho = 0.498$, $p = 2.2 \times 10^{-12}$). Semantic errors showed significant alignment ($\rho = 0.322$, $p = 1.5 \times 10^{-5}$), as did mixed ($\rho = 0.407$, $p = 2.1 \times 10^{-8}$), unrelated ($\rho = 0.435$, $p = 2.6 \times 10^{-9}$), nonword ($\rho = 0.347$, $p = 3.2 \times 10^{-6}$), and no-response ($\rho = 0.419$, $p = 5.1 \times 10^{-9}$). All survived permutation testing ($p_{\text{perm}} < 0.001$). The per-item cosine similarity distribution was strongly left-skewed: 77.1% of items had cosine similarity above 0.90, with median 0.937. This alignment is perturbation-induced rather than a baseline property of the model: at zero modification the model–patient correlation is near zero ($\rho = 0.134$), rising to a plateau of roughly 0.50–0.55 under moderate-to-strong perturbation and peaking at $\rho = 0.570$ (noise 1.8, 90% modification).

The per-patient best-match analysis provides a complementary, individual-level view. Correlating each patient's 175-item correct-rate vector against all perturbation-condition vectors (an 88 × grid matrix computed by vectorized z-score multiplication) yielded a mean best-match r of 0.355, 15.6 standard deviations above the permutation null, with optimal parameters clustering around Layer 19.7, noise 1.54, and modification 66.7%—moderate perturbation at semantic processing layers. This clustering is not trivially expected: under an artifactual alignment the optimal parameters would be uniformly distributed across the grid. The most divergent items were short, common words (e.g., ball, iron, rake) that patients name correctly but that LLaVA misidentifies visually, producing unrelated responses—visual-recognition failures in the model rather than language-processing failures, whose exclusion would raise the mean cosine similarity further.

A held-out test corroborates the correspondence: ten-fold cross-validated prediction of patient per-item difficulty from the model's item-level features yielded r = 0.303 (RMSE = 0.095), exceeding a permutation null (p = 0.04). The alignment therefore generalizes to unseen items rather than reflecting overfitting to the full 175-item set.

| Category | Pearson r | p | Spearman ρ | p | Sig. | 95% CI | Seeds | Stage |
|---|---|---|---|---|---|---|---|---|
| correct | +0.333 | 6.8e-06 | +0.498 | 2.4e-12 | *** | [+.374,+.605] | 100% | Sem-lex |
| semantic | +0.195 | 9.8e-03 | +0.320 | 1.6e-05 | ** | [+.172,+.456] | 100% | Sem-lex |
| mixed | +0.394 | 6.6e-08 | +0.405 | 2.8e-08 | *** | [+.266,+.524] | 100% | Sem-lex |
| unrelated | +0.201 | 7.6e-03 | +0.435 | 1.8e-09 | *** | [+.297,+.559] | 100% | Sem-lex |
| nonword | +0.354 | 1.6e-06 | +0.345 | 2.9e-06 | *** | [+.205,+.471] | 100% | Partial |
| noresponse | +0.232 | 2.0e-03 | +0.420 | 6.9e-09 | ** | [+.290,+.539] | 100% | Failure |
| formal | +0.012 | 8.7e-01 | +0.112 | 1.4e-01 | n.s. | [-.041,+.256] | 0% | Phon. |
| neologism | +0.057 | 4.6e-01 | +0.084 | 2.7e-01 | n.s. | [-.058,+.227] | 1.2% | Phon. |

***Table 1.*** *Per-category item-level correlation between POLAR patient error rates and LLaVA perturbation error rates across 175 PNT items. Each row correlates one category's per-item rate in patients against the same category's per-item rate in the model, so a positive value means the items on which patients produce that error are the items on which the model produces it. Columns: Pearson r and Spearman ρ with their two-sided p-values; Sig., significance of the Spearman test (*** p < 0.001, ** p < 0.01, * p < 0.05, n.s. not significant); 95% CI, percentile bootstrap interval for ρ from 10,000 resamples of the 175 items; Seeds, the percentage of the 50 random seeds whose individual correlation reached p < 0.05, which indicates whether a result rests on the aggregate or holds within single runs; Stage, the processing stage the category is assigned to in two-step models of word production — Sem-lex, semantic-lexical; Phon., phonological; Partial, arising at either stage; Failure, no response produced. The stage assignment is theoretical, taken from the model of production the predictions derive from, and is not estimated from these data. The two phonological categories are the ones whose intervals include zero.*

Per-patient best-match analysis showed that each patient's error profile was best matched by a specific perturbation condition, with best-match correlation inversely related to aphasia severity ($\rho = -0.741$ with WAB-AQ)—more severe patients were matched by more perturbed conditions (Figs. 5–6). Among the 175 items, the hardest for patients (stethoscope, microscope, pretzel) were also among the hardest for LLaVA.

Among the 175 items, the hardest for patients—stethoscope, microscope, binoculars, thermometer, slippers—were difficult for the model as well, with the instructive exception of slippers, which LLaVA almost never names correctly because of visual ambiguity rather than linguistic difficulty. Divergent items are themselves informative: they are typically items with idiosyncratic visual properties (where the model sees a generic shape and produces unrelated responses while patients recognize the object easily) or items with high phonological complexity but clear visual identity (where patients struggle with phonological assembly while the model identifies the object).

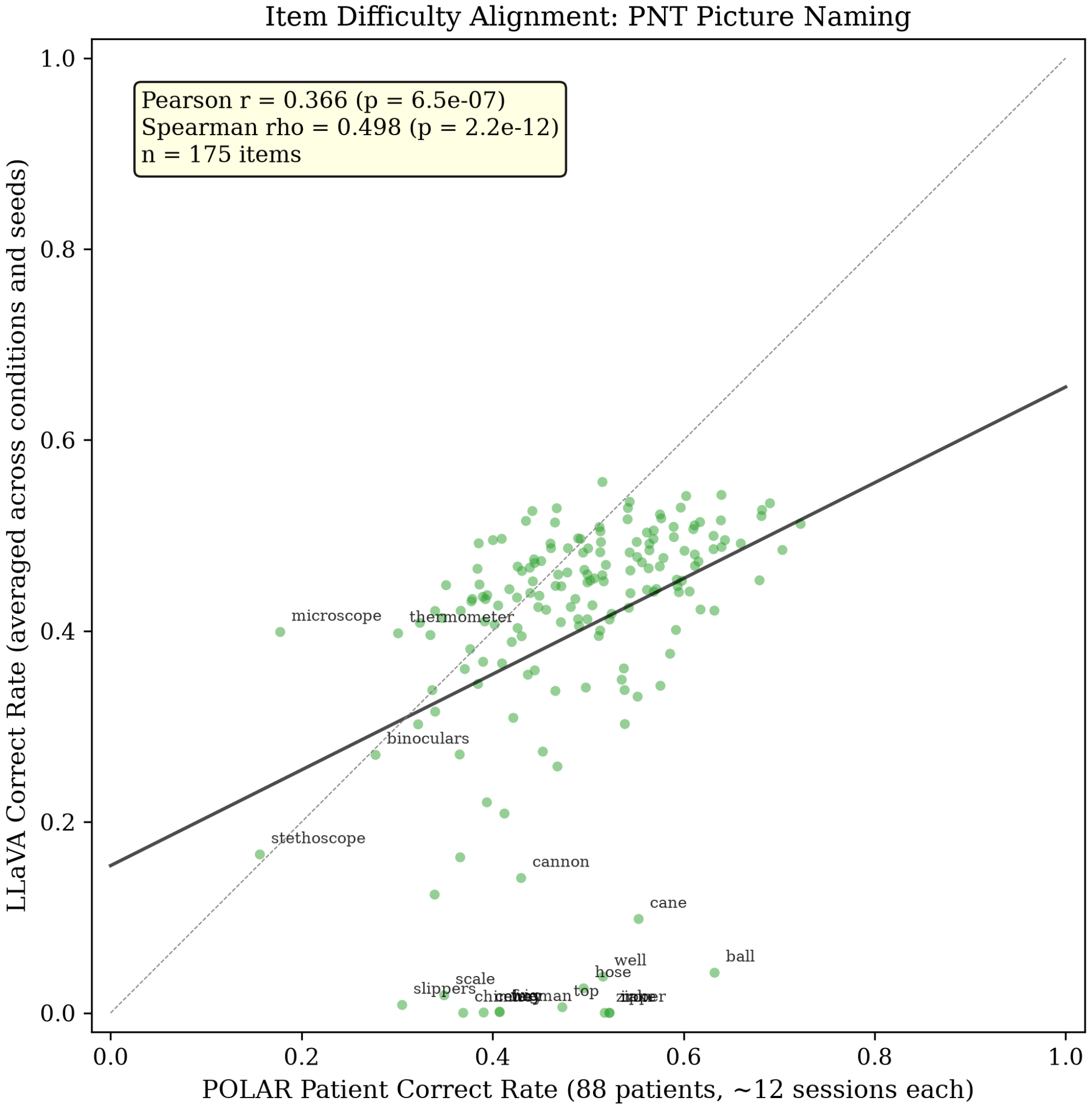

***Figure 5.*** *Item difficulty alignment. POLAR patient correct rate versus LLaVA correct rate for each of 175 PNT items. Spearman* $\rho = 0.498$ *(*$p = 2.2 \times 10^{-12}$*). Outlier items labeled. Diagonal: perfect correspondence; regression line in red.*

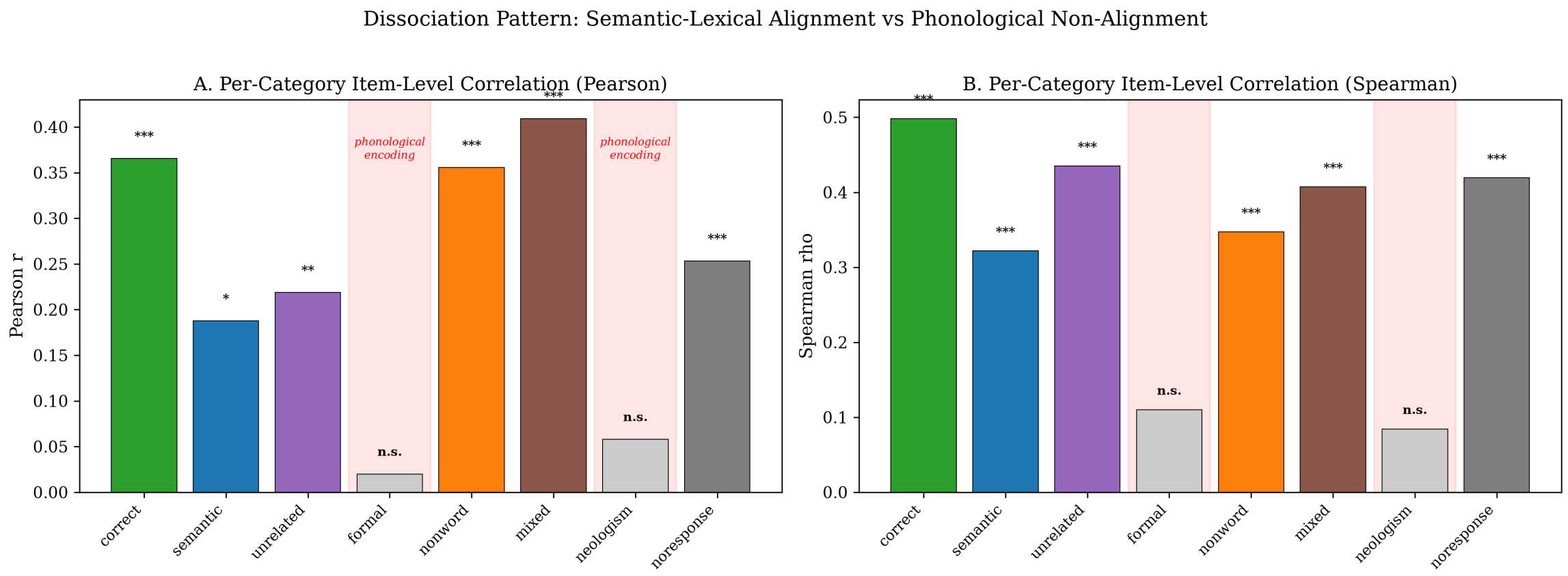

***Figure 6.*** *Dissociation summary: correlation strength by error category. Green/blue: significant categories (Spearman ρ, Pearson r). Red/orange: non-significant (formal, neologism). Significance stars annotated. The dissociation boundary falls between semantic-lexical and phonological processing stages.*

## 3.4 Phonological error categories show no alignment

Formal errors ($\rho = 0.112$, $p = 0.14$) and neologisms ($\rho = 0.084$, $p = 0.27$) showed no significant alignment—consistent with the prediction that LLaVA's lack of phonological encoding prevents patient-like formal error patterns. Per-seed stability was remarkable: the 50 individual-seed ρ values ranged from 0.488 to 0.508 (mean 0.496, SD 0.0042); split-half reliability reached 0.9998 (Spearman–Brown corrected). Comparing against a priori predictions: Group A confirmed (all $p < 10^{-5}$), Group B confirmed (both n.s.), Group C resolved—nonword aligned ($\rho = 0.347$, placing it with Group A) and no-response aligned ($\rho = 0.419$). The alignment of nonword errors despite their phonological definition is discussed in Section 4.2.

The per-seed analysis underscores that the dissociation is architectural rather than stochastic. For correct responses the 50 per-seed Spearman ρ values had a coefficient of variation of only about 0.85%, meaning LLaVA's item-difficulty ranking is essentially a deterministic property of the architecture interacting with the PNT stimuli, not an artifact of any noise realization. By contrast, 0/50 seeds reached significance for formal errors and only 1/50 for neologisms, versus 50/50 for every aligned category, and split-half reliability reached 0.9998 (Spearman–Brown corrected) throughout—confirming a genuine absence of signal, not measurement instability. Crucially, no alternative grouping fits: semantic errors (base rate 2.7%, the rarest aligned category) align significantly while formal errors (base rate 6.3%, more than twice as frequent) do not, ruling out a base-rate-driven artifact. The boundary is defined by processing requirements, not by the statistical properties of the categories.

## 3.5 The dissociation is layer-specific

Layer-stratified analysis revealed an inverted-U alignment profile for the six aligned categories (Fig. 7), peaking at L22 (55% depth) with $\rho = 0.514$ and declining at both shallower and deeper layers. The two non-aligned categories remained near zero across all layers, forming a clearly separable lower band. The minimum gap between the lower 95% CI of the semantic-lexical band and the upper 95% CI of the phonological band is 0.09 ρ units at L22, remaining positive across L18–L28—a robust, statistically demonstrable dissociation.

The separation can be quantified across the full depth: the minimum gap between the two bands (lower 95% CI of the semantic-lexical group minus upper 95% CI of the phonological group) was about 0.07 in the early layers and exceeded 0.15 from L13 onward, reaching roughly 0.30 at the L22 peak. This sustained separation across all 40 layers, computed from 50 independent seeds with uncertainty quantification, is the single strongest piece of evidence for the architectural dissociation.

Correct-rate alignment peaks sharply at L22, the layer used for the dual-stream analysis. The aligned categories peak across a broad mid-to-late band rather than a single locus (mixed L21, unrelated L15, no-response L16, semantic L35, nonword L39), while the two non-aligned phonological categories peak only weakly (formal L19, $\rho = 0.27$; neologism L8, $\rho = 0.17$). The identification of L22 as a semantic processing layer is grounded in probing studies showing that middle-to-late transformer layers encode semantic rather than syntactic information (Tenney et al., 2019; Hewitt & Manning, 2019).

Identifying L22 as a semantic processing layer is grounded in probing studies of transformer architectures (Tenney et al., 2019; Rogers et al., 2020), which find that early layers encode surface and syntactic features, middle layers encode semantic and relational content, and late layers specialize for output prediction. In LLaVA, where visual features are projected into the language model's embedding space, the mid-layers are where visual-semantic integration matures—the model has formed a stable semantic representation of the depicted object but has not yet committed to an output token. The peak at the 55th percentile of depth therefore corresponds to the range where semantic-lexical representations are maximally informative, exactly where alignment arising from shared semantic-lexical processing should appear.

A layer-axis result of the same shape has been reported independently. Guan et al. (2026) applied subtraction analysis to the same model, treating each perturbation seed as a subject and testing along the layer axis with threshold-free cluster enhancement, and recovered a deep-layer cluster favouring phonemic over semantic errors that replicated on held-out splits; the semantic-favouring direction was a consistently signed but non-significant trend. The two analyses ask different questions — theirs where in the model each error class is generated, ours which classes align with patients — and use different contrasts, so they are not redundant. That both single out the phonological categories, and both locate the relevant structure in the deeper half of the network, is convergence from independent directions.

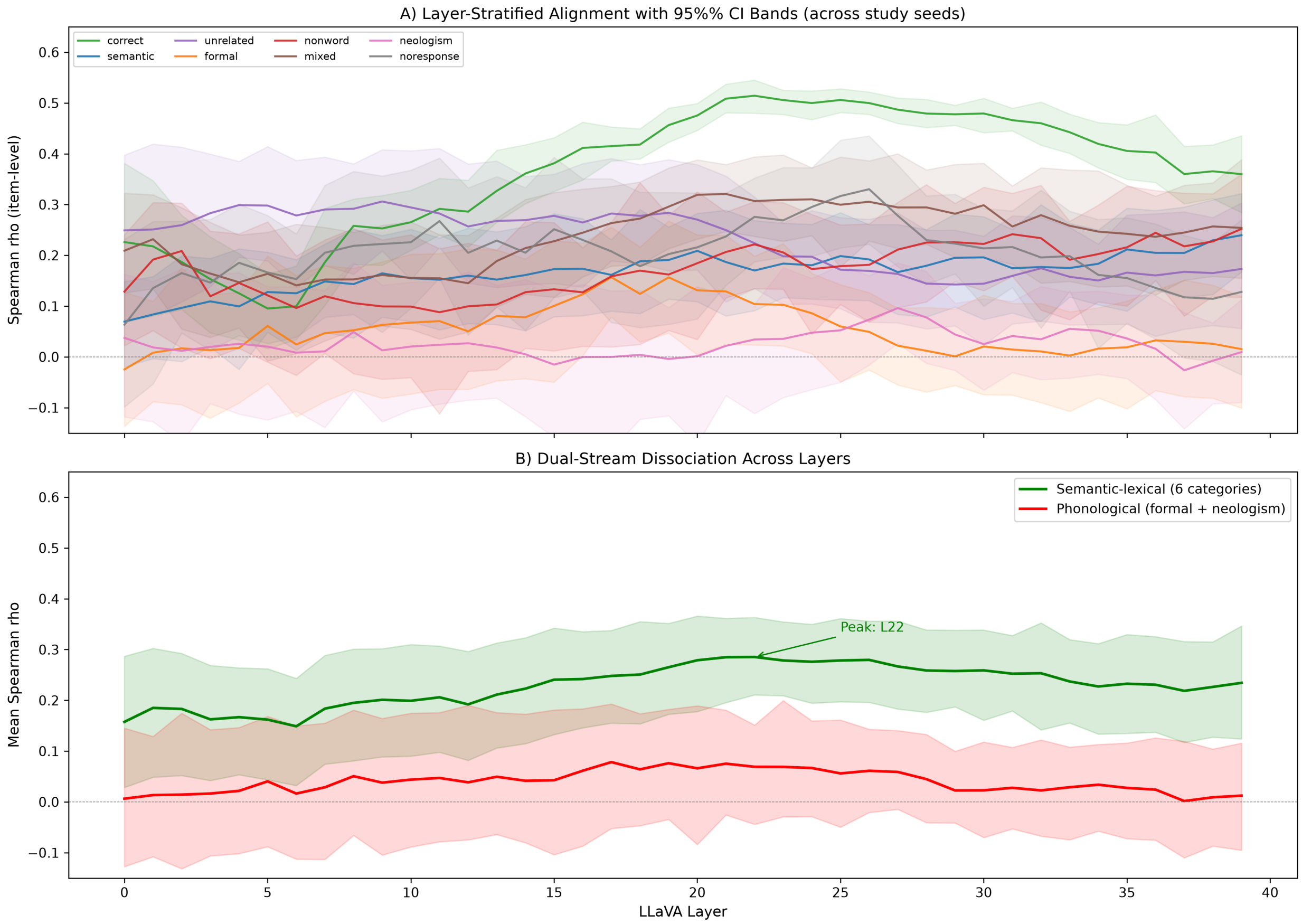

***Figure 7.*** *Layer-stratified alignment with 95% CI bands from 50 seeds. (A) All eight categories across 40 layers. Correct (green) shows inverted-U peaking at L22; formal (orange) and neologism (pink) stay near zero. (B) Dual-stream dissociation: semantic-lexical mean (green band) versus phonological mean (red band), fully separated at every layer. This is the core evidence for architectural dissociation.*

### 3.6 Robustness and error profiles

Stratification by WAB-AQ severity tertiles showed significant alignment in all groups: severe ($\rho$ = 0.421), moderate ($\rho$ = 0.500), and mild ($\rho$ = 0.432; Fig. 8; all $p < 0.001$). Split-half reliability exceeded r = 0.998. Per-item cosine similarity averaged 0.882 (Fig. 9), though dominant error concordance was 0% (excluding no-response): patients' most frequent error type never matched LLaVA's—predicted by the architectural dissociation, since patients produce phonological errors that LLaVA cannot. Syllable count correlated negatively with cosine similarity ($\rho$ = −0.309, p = $3.1 \times 10^{-5}$): longer words showed greater profile divergence, consistent with more opportunities for phonological errors. Additional robustness analyses are in SI Figures S6–S12.

WAB-AQ was uncorrelated with alignment strength ($\rho$ = −0.025, p = 0.83), confirming that the correlation is not a trivial consequence of shared easy-item success. The per-patient best-match quality, however, was strongly severity-dependent: WAB-AQ correlated with best-match r at $\rho$ = −0.750. Severe patients could not be well matched by any perturbation condition because their errors are dominated by phonological types the model cannot generate, whereas mild patients—whose errors are predominantly semantic and no-response—were well reproduced. This severity-matchability gradient is itself evidence for the dual-stream interpretation: if the model merely lacked random capability, matchability would fall uniformly with severity; the fact that it falls specifically because of phonological errors identifies phonological processing as the missing architectural component. WAB-AQ also tracked optimal noise ($\rho$ = −0.282), so more impaired patients matched higher-noise conditions, consistent with perturbation-as-impairment.

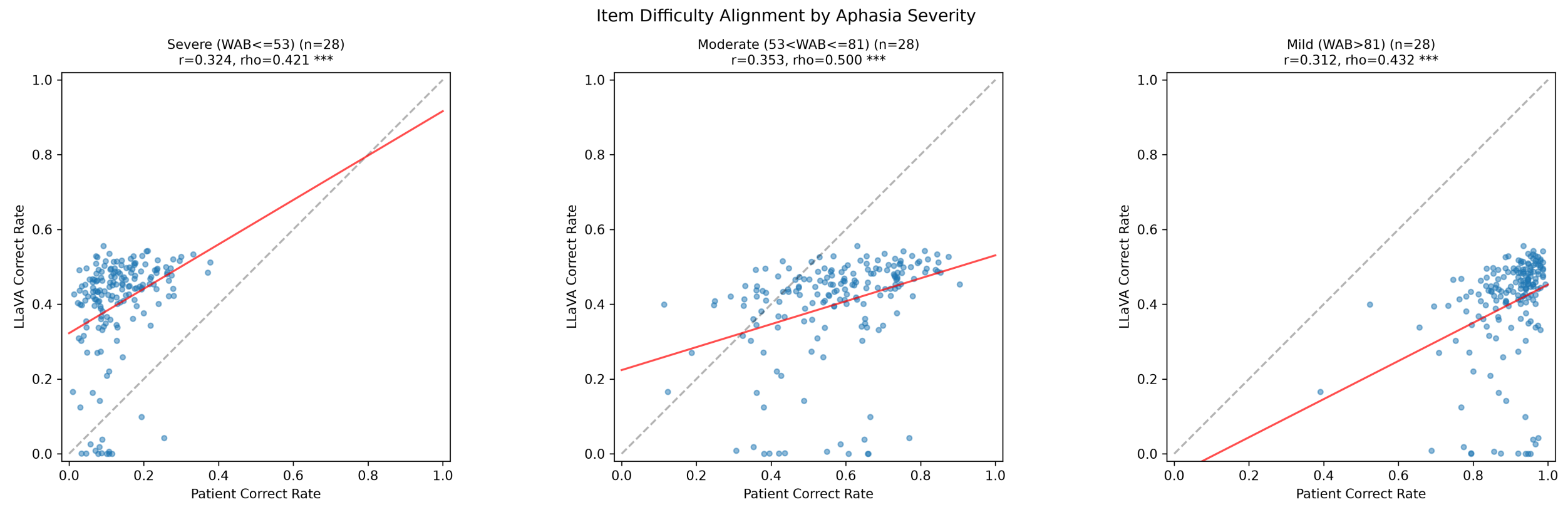

***Figure 8.*** *Severity-stratified alignment. Per-item correct rate scatter for three WAB-AQ tertiles (severe/moderate/mild, n = 28 each). All groups show significant alignment (p < 0.001). The dissociation pattern is preserved across severity levels.*

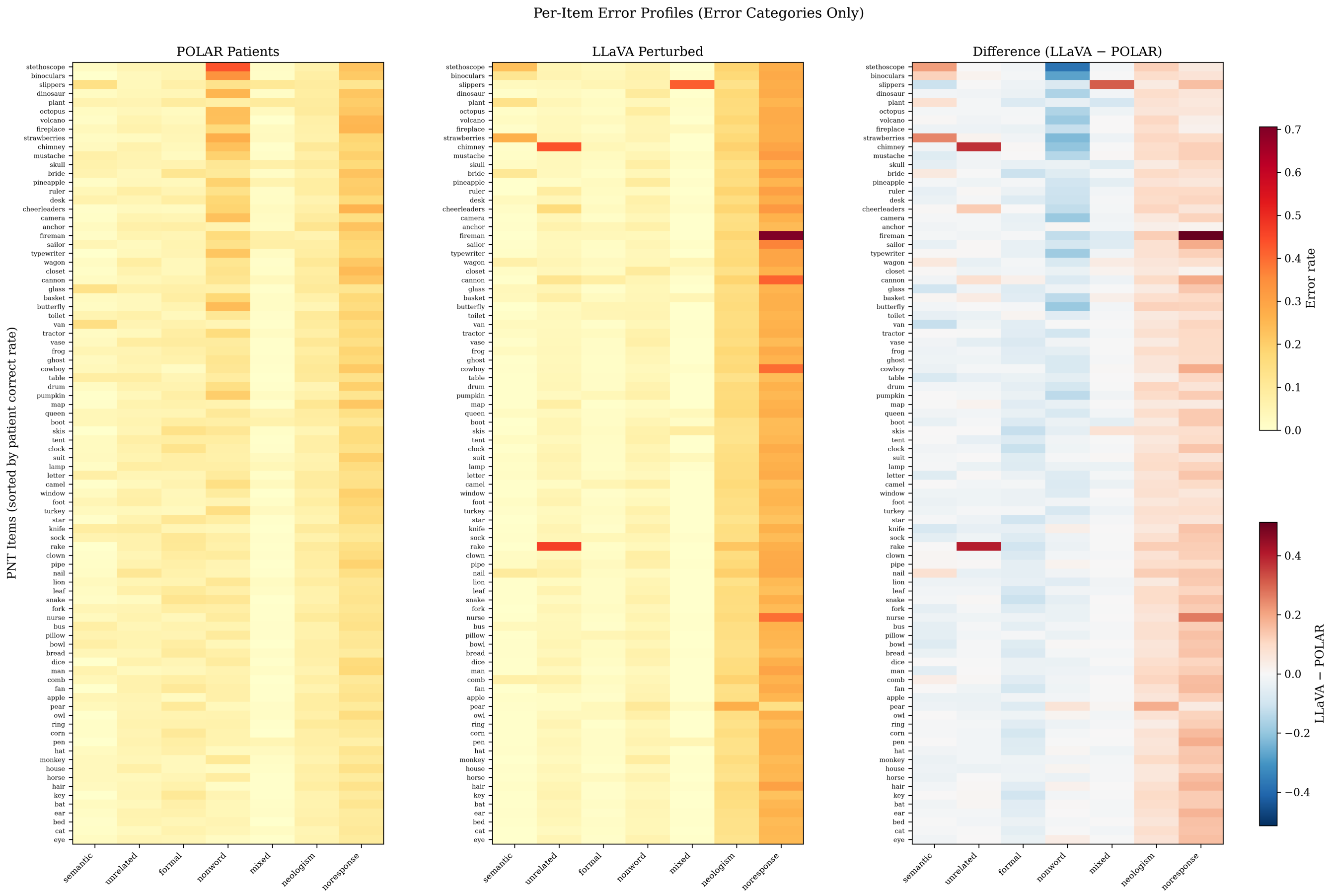

***Figure 9.*** *Side-by-side error profile heatmaps. (A) POLAR patient per-item profiles. (B) LLaVA per-item profiles. (C) Difference. Items sorted by patient difficulty (hardest at top). Overall shapes are similar, but the specific error composition diverges for difficult items: patients produce nonword/formal errors; LLaVA produces noresponse/neologism.*

This pattern—shape-similar profiles with divergent dominant types—has a clear two-stage interpretation. The high cosine similarity reflects agreement on overall item difficulty: both systems find the same items challenging. The zero hard concordance reflects different failure mechanisms: when patients fail on a difficult item, the failure mode depends on residual phonological encoding capacity (nonword or formal errors if partially intact, neologism if not); when LLaVA fails on the same item, the failure is binary collapse to no-response or unstructured output, because it has no phonological stage to fail partially. The word-length effect provides a within-item version of the same test: difficulty alignment rises with syllable count (short $\rho$ = 0.420, medium 0.463, long 0.662) even as profile similarity falls, because the same long words are hard for both systems but elicit graded phonological degradation in patients and collapse in the model.

### 3.7 Cross-model validation

To test whether the 6/8 alignment requires specific architecture or scale, we applied weight perturbation to four additional 7B VLMs spanning a 2×2 architecture matrix: LLaVA-7B (sequential + dense MHA), Molmo-7B-O (parallel + dense MHA),

Molmo-7B-D (sequential + GQA), and Qwen2-VL (sequential + GQA). Each was run on the same perturbation grid and the same picture-naming task, and each produced item-level alignment on correct rate of comparable magnitude to the reference model ($\rho = 0.21$–$0.27$, all $p < 0.01$), so perturbation reproduces the difficulty structure of the task across architectures.

Whether the per-category dissociation itself depends on scale or on attention design is a separate question, and one these runs cannot settle: the reference figures and the cross-model figures come from different inference implementations, so a difference between them is not attributable to the models alone. Answering it requires a comparison inside a single implementation — LLaVA-7B against LLaVA-13B for scale, and a model matched to LLaVA-7B on vocabulary, depth and width but differing in attention design for architecture — which is reported separately. What the present set establishes is that the correct-rate correspondence is not peculiar to one model.

Perturbation models static capacity, not recovery: among 68 patients with ≥4 sessions, accuracy slope versus noise slope showed no correlation ($\rho = -0.223$, $p = 0.067$), indicating that instantaneous weight perturbation does not capture the dynamic trajectory of aphasia recovery.

Two caveats temper the negative longitudinal finding. The POLAR sessions span a relatively short treatment window over which accuracy changes are modest, so the small session-to-session shifts may be insufficient to register in perturbation-parameter space even if recovery does correspond to reduced perturbation; an acute-to-chronic comparison with larger changes might detect it. In addition, the two treatment arms target different stages (phonological versus semantic), so recovery patterns may differ by arm in ways masked by the aggregate analysis. We therefore read the result as a boundary on the current design rather than a refutation of the perturbation-recovery hypothesis: in computational terms, perturbation operates on the data flow while recovery operates on the model itself, mirroring the neurobiological distinction between acute functional disruption and chronic reorganization.

## 4. Discussion

### 4.1 Computational architectural dissociation as a method

The central methodological contribution is computational architectural dissociation: using known architectural differences between artificial and biological systems to derive testable predictions about cognitive organization. Unlike classical double-dissociation studies, this approach requires only a model with a known architectural absence and a clinical dataset with sufficient scale. The method is generalizable: any cognitive function operationalizable as a task with item-level variation can be tested. A key distinction from brain–AI alignment deserves emphasis: studies showing language model representations correlate with neural activity demonstrate representational similarity but cannot address whether processing stages are separable. Our approach uses architectural absence—what a model lacks—to make positive predictions about cognitive architecture.

The method is generalizable beyond word production. Any cognitive function operationalizable as a task performable by both humans and an AI system is amenable: identify what the AI lacks, predict which behavioral categories will and will not align, and test empirically. It is most powerful when the AI lacks a specific, identifiable component rather than differing diffusely. Candidate applications include testing the modularity of syntactic processing using models without hierarchical structure, the separability of orthographic and phonological reading routes using text-only versus speech-trained models, and the independence of pragmatic reasoning from literal comprehension using models with and without theory-of-mind capabilities. In each case the inferential role of the architectural absence is the same as that of a controlled lesion or TMS disruption, with the advantage that the removal is exact, verifiable, and reproducible.

### 4.2 Evidence for two-stage word production

The alignment/dissociation boundary maps precisely onto Dell's predicted stage boundary. The six aligned categories are all generable through semantic-lexical competition; the two non-aligned require phonological processing. The dissociation does not fall between easy/hard or high/low frequency categories—only between semantic-stage and phonological-stage types.

The alignment of nonword errors ($\rho = 0.347$) despite their phonological definition is the most informative single finding. In Dell's model, nonword errors arise from perturbation of the phonological stage, yet item-level data show that which items elicit nonwords correlates between patients and LLaVA—a model without phonological processing. This implies vulnerability to nonword errors is determined at the semantic-lexical stage: items with weak or ambiguous semantic representations produce degraded outputs regardless of downstream phonological integrity.

Why does this argument not apply to formal errors? Formal errors require phonological neighborhood access: "mat" for "cat" requires the system to represent that "mat" sounds like "cat." Which items have exploitable phonological neighbors depends on phonological representations that LLaVA lacks. Items eliciting formal errors in patients are determined by phonological neighborhood structure, not semantic difficulty—hence no item-level correlation with LLaVA.

This nonword–formal distinction refines the two-stage model. Dell's model treats both as arising from the phonological stage, but our data show different item-level determinants: nonword errors predicted by semantic-lexical difficulty (shared with LLaVA), formal errors by phonological neighborhood structure (absent from LLaVA). This suggests the phonological stage has internal structure—a distinction between output degradation and neighborhood substitution. Previous Dell model validations have not tested item-level patterns within the phonological stage (Schwartz et al., 2006), making this refinement empirically novel.

The interpretation is reinforced by the observation that nonword errors in patients and the model occur on the same items but through different mechanisms: phonologically complex words fail at phonological encoding in patients and at the semantic-lexical level in the model, a common-cause convergence on overall item difficulty rather than shared phonological processing. The per-category peak layers add a finer view of the hierarchy: categories peaking early (unrelated, no-response) reflect failure to establish even a coarse semantic representation, those peaking at mid layers (mixed, correct) reflect core semantic-lexical processing, and those peaking late (semantic, nonword) reflect output-level competition—a computational analog of Dell's processing stages mapped onto transformer depth.

The spectral analysis of mixed errors further supports two-stage architecture: mixed errors are spectrally much closer to semantic errors (distance 0.281) than formal errors (2.80), consistent with generation through semantic competition with incidental phonological overlap. The sharp correct-rate alignment peak at L22 localizes the dominant semantic-stage contribution to the mid-to-late layers.

The mixed-error finding also speaks to the interactive-versus-serial debate. Dell's interactive model predicts a "mixed-error boost"—mixed errors occurring above the product of independent semantic and phonological error probabilities—because feedback amplifies words that are both semantically and phonologically active (Nozari et al., 2011). LLaVA, which has no feedback mechanism, produces mixed errors at rates consistent with their spectral proximity to semantic errors, not at boosted rates. This does not prove that human mixed errors lack phonological involvement; it proves such involvement is not necessary for their production. If the boost is confirmed in patient data, the contrast would constitute a positive dissociation—patients showing the boost (interactive feedback) and LLaVA not (a mechanism it lacks)—providing a direct future test of the interactive-feedback hypothesis.

### 4.3 What VLMs lack reveals what humans have

The formal error result provides positive evidence that phonological neighborhood access is a distinct processing resource, not emergent from distributional statistics. The layer-specificity adds architectural detail: semantic-lexical alignment peaks at L22 while phonological categories remain flat across all 40 layers, indicating no processing depth approximates phonological computation. This demonstrates that phonological neighborhood competition is not captured by even the largest text-trained language models, requiring dedicated phonological representations.

The formal-error dissociation further suggests that phonological similarity, not orthographic similarity, drives item-specific formal vulnerability. Transformers trained on text develop representations of orthographic similarity, but orthographic and phonological similarity are only partially correlated in English (consider cough and dough, orthographically similar yet phonologically distant). The absence of formal-error alignment therefore implicates the phonological neighborhood structure of the lexicon—a representational system not captured by text statistics—and yields a concrete prediction: augmenting a VLM with an explicit phonological module trained on speech with phonemic representations should make formal-error alignment emerge, while its emergence in a purely text-trained model would challenge the two-stage account. The layer-specific localization also offers a neuroimaging prediction: if the alignment layer maps onto cortex, the corresponding regions should be those implicated in semantic-lexical processing (posterior temporal and inferior parietal cortex; Bonilha et al., 2017; Hickok, 2012) rather than phonological processing in posterior inferior frontal gyrus.

### 4.4 Spectral analysis as a mechanistic interpretability tool

The output probability spectral analysis represents a methodological contribution independent of the clinical application. Rather than probing internal representations, it characterizes what the model outputs—the shape of its probability distribution—providing an external, observable window into internal processing state. Three properties make spectral signatures useful: they are observable without internal model access (enabling cross-architecture comparison), continuous and ordinal (supporting quantitative analysis), and mechanistically linked (the spectral hierarchy maps directly onto perturbation's effect on internal competition). This contribution extends to adversarial robustness, model compression, and clinical NLP.

Three properties make spectral signatures useful as interpretable objects. First, they are architecture-independent: spectral flatness, KL divergence, and the other observables are defined for any probability distribution and so apply to any current or future model that produces output distributions, enabling cross-architecture comparison without internal access. Second, they are compositional: the spectral distance between two categories has a direct interpretation as the dissimilarity of the processing states that produce them, supporting clustering, hierarchy, and distance-prediction analyses that classifier labels alone cannot. Third, they are clinically translatable: the hierarchy from peaked (correct) to flat (neologism) maps onto the clinical severity continuum, providing a quantitative bridge between model behavior and clinical assessment that attention-pattern-based interpretability does not offer.

### 4.5 Limitations, future directions, and conclusion

The negative longitudinal result indicates that perturbation models static capacity, not recovery dynamics. All analyses are PNT-specific; generalization requires further validation. Cross-model experiments establish that 13B-scale capacity is necessary, but single seeds per 7B model limit statistical power for the negative finding.

Future directions include testing 13B+ models with diverse architectures, integration with the ARC neuroimaging dataset (Bonilha et al., 2019) for lesion-layer mapping, extension to other language tasks, and augmenting a VLM with explicit phonological encoding—the most direct test of the two-stage hypothesis.

Taken together, the approach yields findings at three levels of granularity. At the coarsest, it confirms that semantic-lexical retrieval and phonological encoding are architecturally separable stages. At a finer level, it shows that phonological errors are not monolithic: nonword vulnerability tracks semantic-lexical difficulty (and so aligns with the model) while formal vulnerability depends on phonological neighborhood structure (and so does not)—a distinction that Dell's single phonological parameter cannot capture. At the finest, the spectral characterization of output distributions introduces a class of interpretable objects bridging model-internal states and clinical error categories, applicable to any language model regardless of architecture.

In summary, the computational architectural dissociation approach provides independent evidence that human word production involves at least two functionally dissociable stages by demonstrating that a VLM performing semantic competition but lacking phonological processing produces patient-like errors for exactly those categories not requiring phonological access. The convergence of spectral mechanism, item-level validation, predicted dissociation, layer-specific localization, and cross-model capacity dependence across 88 PWA, 175 items, 50 seeds, and 5 VLM architectures constitutes a multi-level evidence package that no single analysis could provide alone.

## Author contributions

Y.Y.: conceptualization, methodology, software, validation, formal analysis, investigation, writing – original draft, writing – review & editing, visualization; R.N.-N.: data curation, writing – review & editing; X.G.: software; S.A.: data curation; R.W.: writing – review & editing; N.S.: writing – review & editing; K.W.: writing – review & editing; S.A.-P.: methodology, data curation, writing – review & editing; S.N.: investigation, resources, writing – review & editing, supervision, project administration; R.H.D.: methodology, writing – review & editing, supervision, project administration; L.B.: data curation, writing – review & editing; C.R.: data curation, writing – review & editing; J.F.: writing – review & editing, supervision, project administration, funding acquisition.

## Acknowledgements

We thank the patients who participated in this research and the POLAR research team for data collection. This work was supported by NIH P50 DC014664 (PI: J. Fridriksson). Computational resources were provided by the Department of Computer Science and Engineering, University of South Carolina (S. Nelakuditi).

## Competing interests

The authors declare no competing interests.

## Data and code availability

The analysis code and the per-seed classification grids and spectral observables it reproduces from will be released on publication, at https://github.com/StevenYY01/Deep_PNT and https://doi.org/10.5281/zenodo.22064376 respectively; between them they regenerate every figure, table and reported number. Access during review is available on request. The Philadelphia Naming Test stimuli are copyrighted and are not redistributed. The POLAR naming data comes from people with aphasia and is held under a data use agreement; the formats of both are documented in the repository so that the pipeline can be run against equivalent material, and the analyses requiring patient data fail with a stated error in its absence rather than substituting anything for it.

## Use of generative AI

The research question, the theoretical framing, the experimental design, the choice of datasets and models, the analysis strategy and the scientific argument of this paper are the authors' own. Generative AI tools were used during implementation: writing and refactoring analysis code, preparing the code and data for release, and drafting and editing portions of the text. They were also used to audit that code, and several defects were found that way; the audit scripts are included in the released repository. No result, figure, table or reported number was produced by a generative model. Every value in this paper is computed by the archived code from the archived data and can be regenerated by running it. The authors reviewed all AI-assisted output and take full responsibility for the content.

# Supplementary Information

## Supplementary Methods

This section provides the extended methodological detail referenced from the main text: the patient cohort and inclusion criteria (Section 2.1), the unit-level weight perturbation procedure and its rationale (Section 2.2), and the formal definitions of the six spectral observables (Section 2.3). Supplementary Note 1 gives the session-weighting formulas, and Supplementary Note 3 compares the perturbation scheme with alternatives.

### SI Methods 1. Patient cohort and inclusion criteria

Participants were 88 patients with aphasia (PWA; chronic post-stroke, >6 months) enrolled in the POLAR (Predicting Outcomes for Language Rehabilitation) study at the University of South Carolina Center for the Study of Aphasia Recovery (C-STAR; NIH/NIDCD P50 DC014664; Fridriksson et al., 2018). Each completed the 175-item Philadelphia Naming Test (PNT) across multiple sessions as part of longitudinal treatment studies, yielding 1,007 valid sessions (median 12 per patient) and 165,375 scored item records. Western Aphasia Battery Aphasia Quotient (WAB-AQ) scores spanned the full severity range (11.2–93.4, mean 60.0). All participants were at least six months post-stroke. Demographic and lesion data are available through the ARC dataset on OpenNeuro (ds004884; Gibson et al., 2024).

Inclusion required a single left-hemisphere stroke documented by CT or MRI, English as the primary language, and sufficient hearing and vision to perform confrontation naming. The cohort spanned the major WAB-classified aphasia subtypes—Broca's, Wernicke's, conduction, anomic, global, and transcortical, with a small subset classified as non-aphasic on the WAB—so that the reported alignment patterns reflect the full spectrum of post-stroke language impairment rather than any single profile. The longitudinal design provides both cross-sectional power (88 PWA × 175 items = 15,400 patient-item observations per session) and the recovery trajectories used in the longitudinal analysis of Section 3.7.

The POLAR study examined predictors of treatment outcome. Participants were enrolled in one of two treatment arms: Phonological Components Analysis (PCA), which targets phonological processing through systematic analysis of a word's phonological features, and Verb Network Strengthening Treatment (VNeST), which targets semantic processing through verb-argument structure. Treatment sessions were interleaved with PNT assessments, providing naming data at multiple time points. Because patients contributed unequal numbers of sessions, per-item error rates were computed as session-weighted averages across patients (formulas in Supplementary Note 1). To verify that the weighting does not distort the item-difficulty structure on which the alignment analysis depends, we compared the session-weighted item-rate vector with the unweighted (per-patient-mean) vector across all 175 items; the two are highly concordant (Pearson $r > 0.99$ for all eight categories; see Supplementary Note 1), confirming that session weighting changes the relative ordering of item difficulty only negligibly.

### SI Methods 2. Unit-level weight perturbation

We used LLaVA 1.6 (Liu et al., 2024) with the Vicuna-13B backbone (Zheng et al., 2023)—a 40-layer transformer with dense multi-head attention (40 query and key/value heads) and a sequential attention-then-MLP layout—paired with a CLIP ViT-L/14 visual encoder (Radford et al., 2021). Each PNT image was presented with the prompt "What is the object in this picture? Answer in one word." Perturbation was applied at the level of weights rather than activations: at a target layer, a fraction of hidden units was selected, and multiplicative Gaussian noise was applied to every aligned weight matrix associated with those units—the query, key, value, and output projections, the MLP weight matrices, and the LayerNorm parameters—according to $w \leftarrow w \times (1 + \varepsilon)$, with $\varepsilon \sim N(0, \sigma)$. Because the perturbation acts on parameters rather than on a forward-pass activation tensor, it is architecture-independent and can be applied identically to any transformer; this property is what enables the cross-model comparison of Supplementary Note 4.

Three parameters varied in full factorial design: target layer (0–39; 40 levels), modification percentage (the fraction of hidden units perturbed; 10–100% in 10 levels), and noise standard deviation σ (1.1–2.0 in 10 levels). The grid was evaluated over the 185 PNT items (175 test items plus 10 warmup items) for 50 hardcoded random seeds, where each seed fixes both which units are selected and the specific noise realization, so the 50 seeds function as independent replications. The full sweep comprises on the order of 37 million naming attempts; all item-level analyses use the 175 test items.

Multiplicative Gaussian weight noise was chosen over alternatives (dropout, weight pruning, activation zeroing) for three reasons. First, it preserves the statistical structure of the weights while reducing their effective signal-to-noise ratio, a plausible computational analog of the uniform connection-strength reduction posited in Dell's account of aphasic naming. Second, it is graded and continuous, allowing fine-grained traversal of the perturbation landscape from near-normal to severely degraded performance, whereas dropout produces discrete, non-monotonic effects. Third, applying it at a specified layer permits the layer-stratified analysis (Section 3.5) that localizes where each error type arises. At the mildest settings ($\sigma = 1.1$, 10% of units) overall correct rate falls 5–10 points below baseline; at the most severe ($\sigma = 2.0$, 100% of units) it falls to 10–25%, dominated by no-response and neologism outputs—spanning the clinical range from mild anomia to severe global aphasia. Alternative interventions (activation patching, direct logit edits) could probe robustness to the perturbation mechanism in future work; see Supplementary Note 3.

One implementation detail is critical. The output probability distribution from which the spectral observables are computed must be captured at the first forward pass through the perturbed network, before the autoregressive generation loop begins. Once generation proceeds, the model's own emitted tokens re-enter as input and overwrite the activation state, silently corrupting the spectral measurements. All spectral analyses reported here use the distribution at the first generated token.

### SI Methods 3. Spectral observables

Six spectral observables—spectral flatness (SF), Zipf exponent (α), participation ratio (PR), Kullback–Leibler divergence from uniform ($D_{KL}$), Shannon entropy (H), and spectral contrast ratio (SCR)—are computed from the top-K (K = 10) output probabilities recorded at the first generated token. They are signal-processing descriptors of the shape of the output distribution, not operator-theoretic spectra. Their formal definitions, the relationship between the spectral map and the error classifier, and the mechanistic reading of each observable are given in Supplementary Note 6.

Because the six observables differ by orders of magnitude in scale—SCR in particular spans a far larger numeric range than SF or α—all spectral distances, including the inter-category distances of Section 3.2 and the eight- versus seven-category comparison, are computed after z-scoring each observable across categories. Without this normalization, Euclidean distances in the six-dimensional spectral space are dominated by SCR alone. Each error category's signature is the vector of these six observables averaged over the responses assigned to that category (Supplementary Fig. S1).

## Supplementary Figures

This section presents 14 supplementary figures organized into three groups that correspond to the three levels of evidence described in the main text. Together with the 9 main text figures, these 23 figures document the complete empirical basis for the computational architectural dissociation. The figures are organized into three groups corresponding to the three evidence levels described in the main text: spectral characterization of the perturbation-induced error taxonomy (Supplementary Figs. S1–S5, supporting main text Section 3.1–3.4 and Figs. 1–3), item-level alignment validation and robustness analyses (Supplementary Figs. S6–S12, supporting main text Section 3.5–3.11 and Figs. 4–8), and longitudinal and layer-transition analyses (Supplementary Figs. S13–S14, supporting main text Section 3.12).

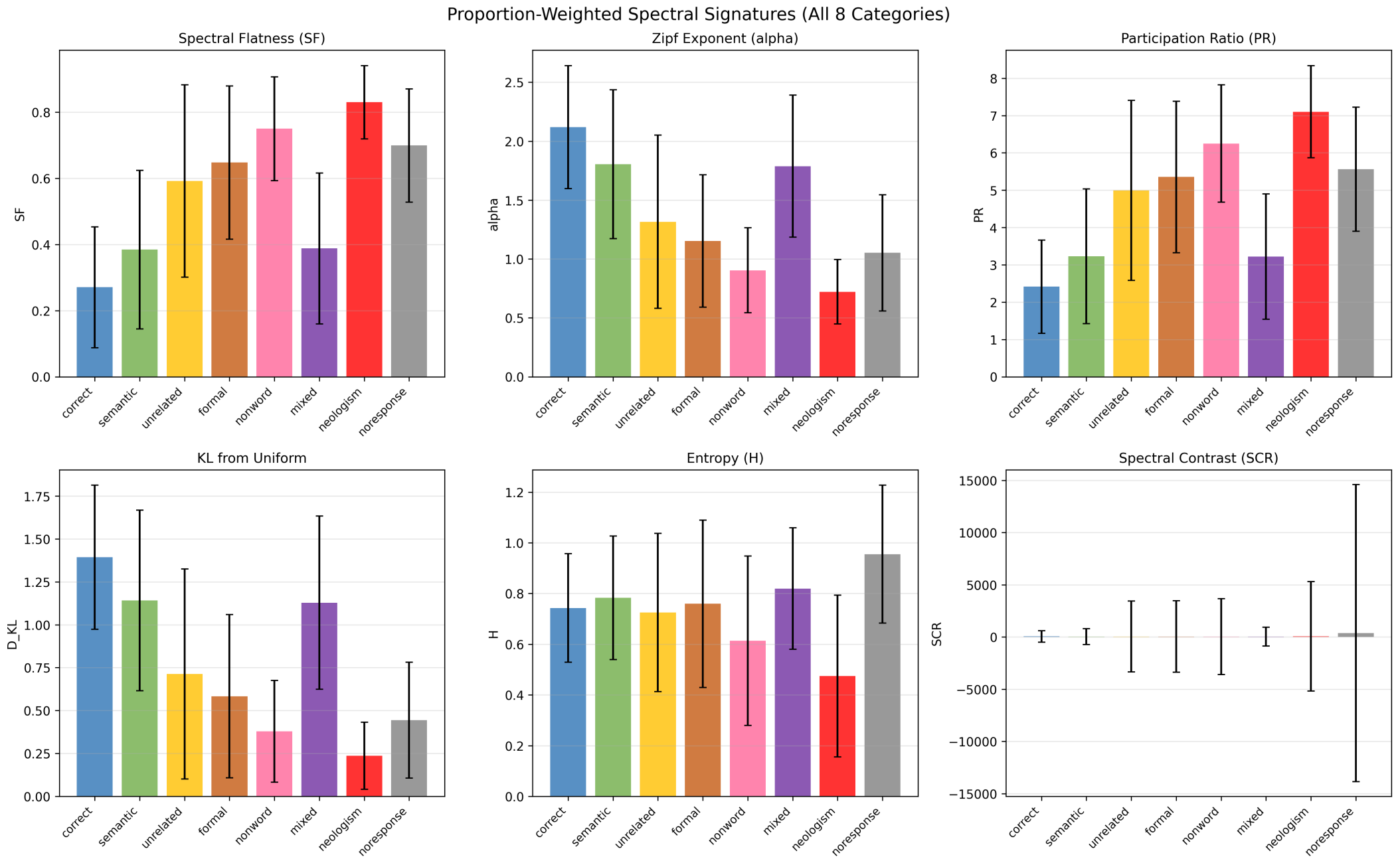

***Supplementary Figure S1.*** *Proportion-weighted spectral signatures for each error category across six observables (spectral flatness, Zipf exponent, participation ratio, KL divergence, prediction entropy, spectral contrast ratio). Each category exhibits a distinct spectral fingerprint. Proportion weighting (multiplying each observable by the category's relative frequency under the given perturbation condition) ensures that rare error types—semantic errors (base rate 2.7%) and mixed errors (1.5%)—are visible despite low base rates. Without proportion weighting, dominant categories (correct, no-response) overwhelm the visualization. The clustering of mixed errors (brown) with semantic errors (blue) rather than with formal errors (orange) is visible across all six observables, supporting the finding that mixed errors are semantic variants in VLMs. Data: 4,000 conditions × 50 seeds, 175 PNT items.*

Spectral characterization of the error taxonomy. The following five figures (S1–S5) provide the full spectral analysis underlying the error taxonomy presented in main text Sections 3.1–3.4 and Figs. 1–3. Where the main text presents the key findings (spectral hierarchy, spectral distance matrix), these supplementary figures provide the underlying data: the complete spectral signatures across all six observables (S1), the dose-response dynamics (S2–S3), the co-occurrence structure validating the spectral distance metric (S4), and the full three-dimensional spectral space (S5).

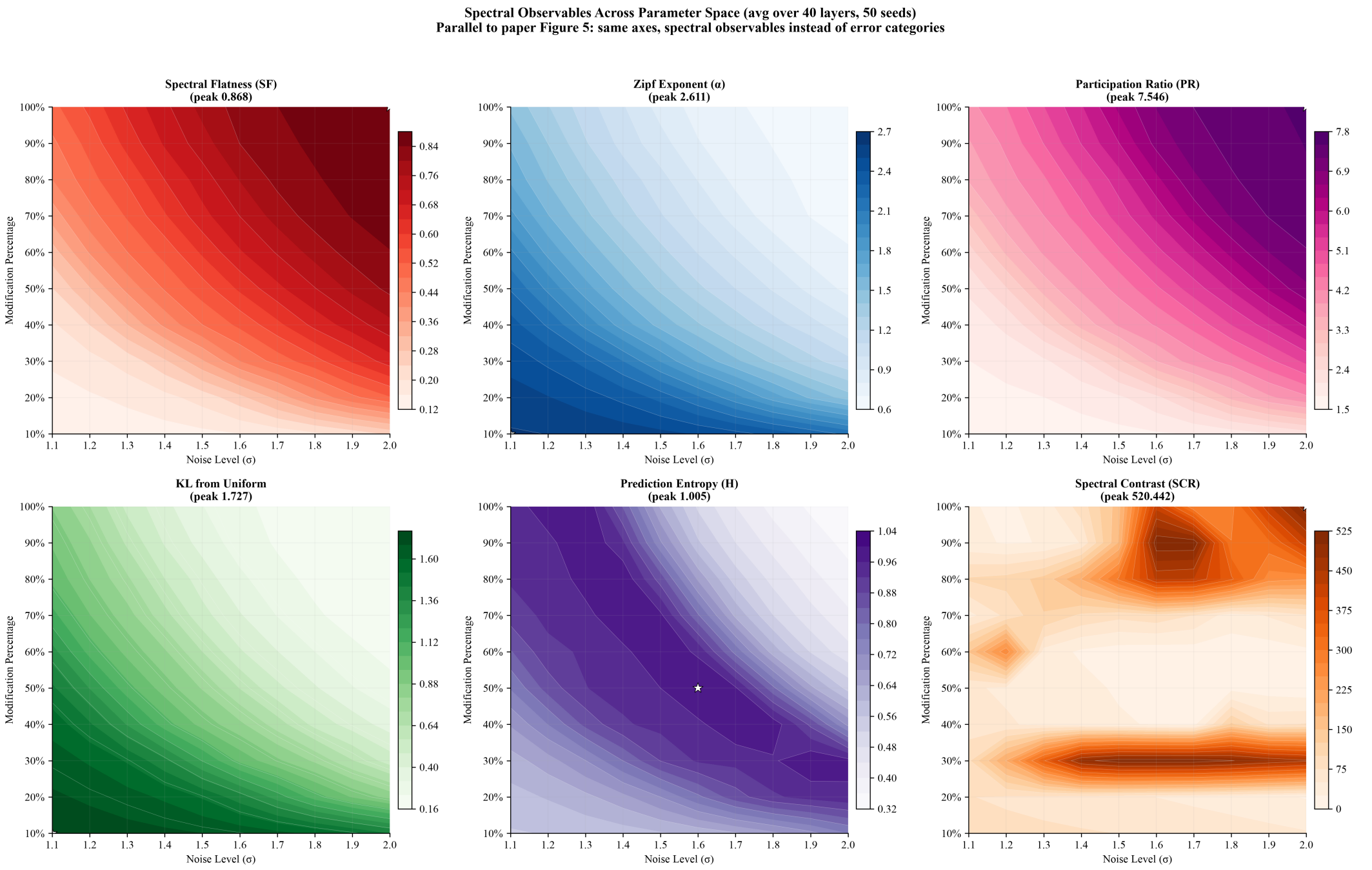

***Supplementary Figure S2.*** *Spectral dose-response curves showing the relationship between perturbation noise magnitude (x-axis, 1.1–2.0 standard deviations) and four key spectral observables across perturbation conditions. (A) Spectral flatness increases sigmoidally with noise. (B) KL divergence from uniform decreases monotonically. (C) Participation ratio increases linearly. (D) Spectral contrast ratio decreases exponentially. Colors indicate different target layers; steeper curves at early layers (blue/purple) reflect greater downstream propagation of perturbation, while shallower curves at late layers (red/orange) reflect more localized effects. All transitions are smooth and continuous, confirming that perturbation-induced error types emerge from graded degradation of the output distribution. Data: 4,000 conditions × 50 seeds.*

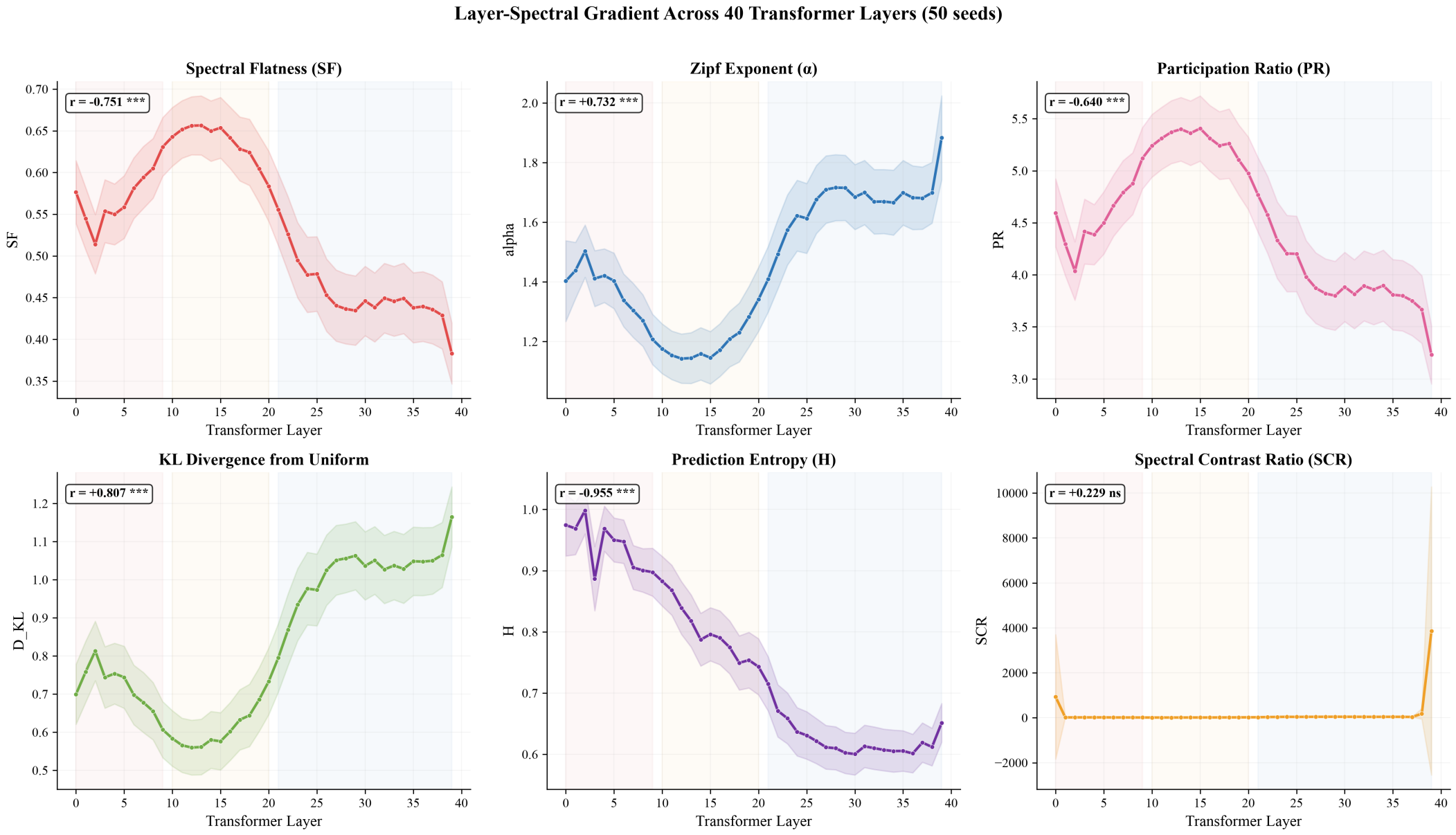

***Supplementary Figure S3.*** *Layer-spectral gradient: spectral flatness as a function of perturbation target layer for different noise levels. Early-layer perturbation (L0–10) produces the strongest spectral changes (largest increase in spectral flatness), consistent with greater downstream propagation of disruption through the model's hierarchical processing. Mid-layer perturbation (L15–25) produces moderate spectral changes concentrated in the semantic processing range. Late-layer perturbation (L30–39) produces the weakest spectral changes, as the disruption affects only the final stages of output generation. This layer-dependent sensitivity profile is consistent with the inverted-U alignment curve observed in the main text (Fig. 6): alignment peaks at the layers where perturbation produces the most informative (neither trivial nor catastrophic) spectral changes.*

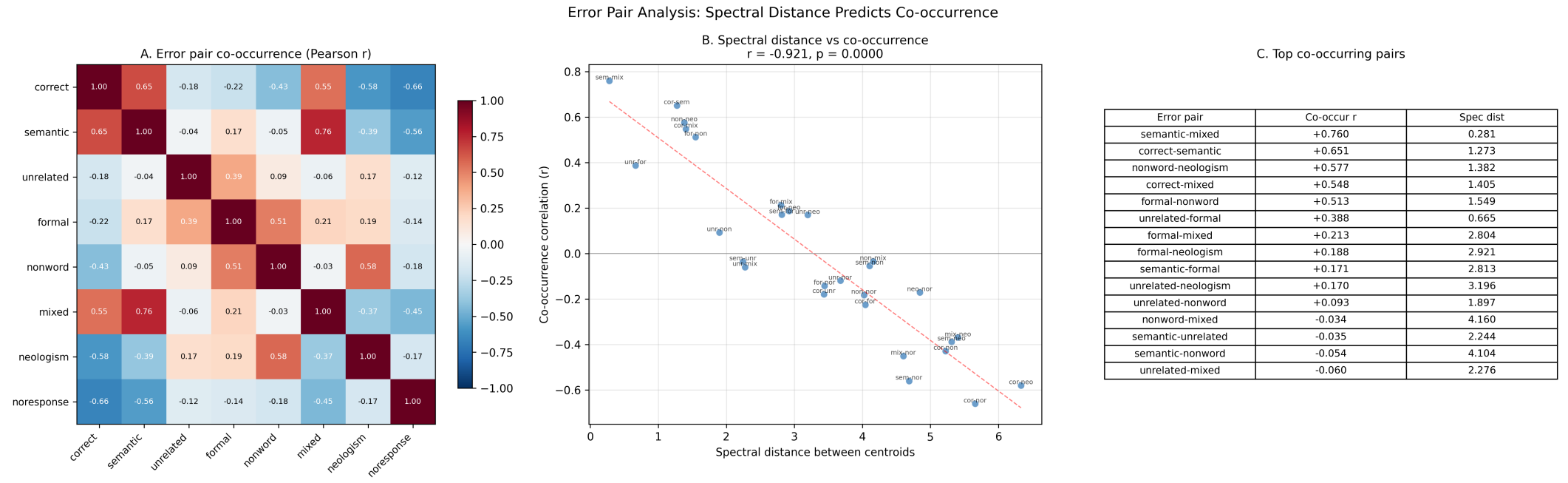

| Error pair | Co-occur r | Spec dist |
|---|---|---|
| semantic-mixed | +0.760 | 0.281 |
| correct-semantic | +0.651 | 1.273 |
| nonword-neologism | +0.577 | 1.382 |
| correct-mixed | +0.548 | 1.405 |
| formal-nonword | +0.513 | 1.549 |
| unrelated-formal | +0.388 | 0.665 |
| formal-mixed | +0.213 | 2.804 |
| formal-neologism | +0.188 | 2.921 |
| semantic-formal | +0.171 | 2.813 |
| unrelated-neologism | +0.170 | 3.196 |
| unrelated-nonword | +0.093 | 1.897 |
| nonword-mixed | -0.034 | 4.160 |
| semantic-unrelated | -0.035 | 2.244 |
| semantic-nonword | -0.054 | 4.104 |
| unrelated-mixed | -0.060 | 2.276 |

***Supplementary Figure S4.*** *Error-pair co-occurrence structure across perturbation conditions. The relationship between the normalized six-dimensional spectral distance of each error-category pair and their empirical co-occurrence is strongly negative (Spearman r = −0.921, p < 0.0001; 85% of variance explained): spectrally proximate error types co-occur more frequently. The mixed–semantic pair has the smallest spectral distance (0.281) and the highest co-occurrence, whereas the formal–mixed pair is widely separated (2.80), supporting the interpretation that mixed errors in the VLM are semantic errors with incidental phonological overlap rather than a distinct dual-channel failure mode. The nonword–neologism pair is spectrally distinct, confirming that the two phonological-error categories occupy separate regions of spectral space. This co-occurrence structure validates the spectral distance metric directly, without reference to any classifier.*

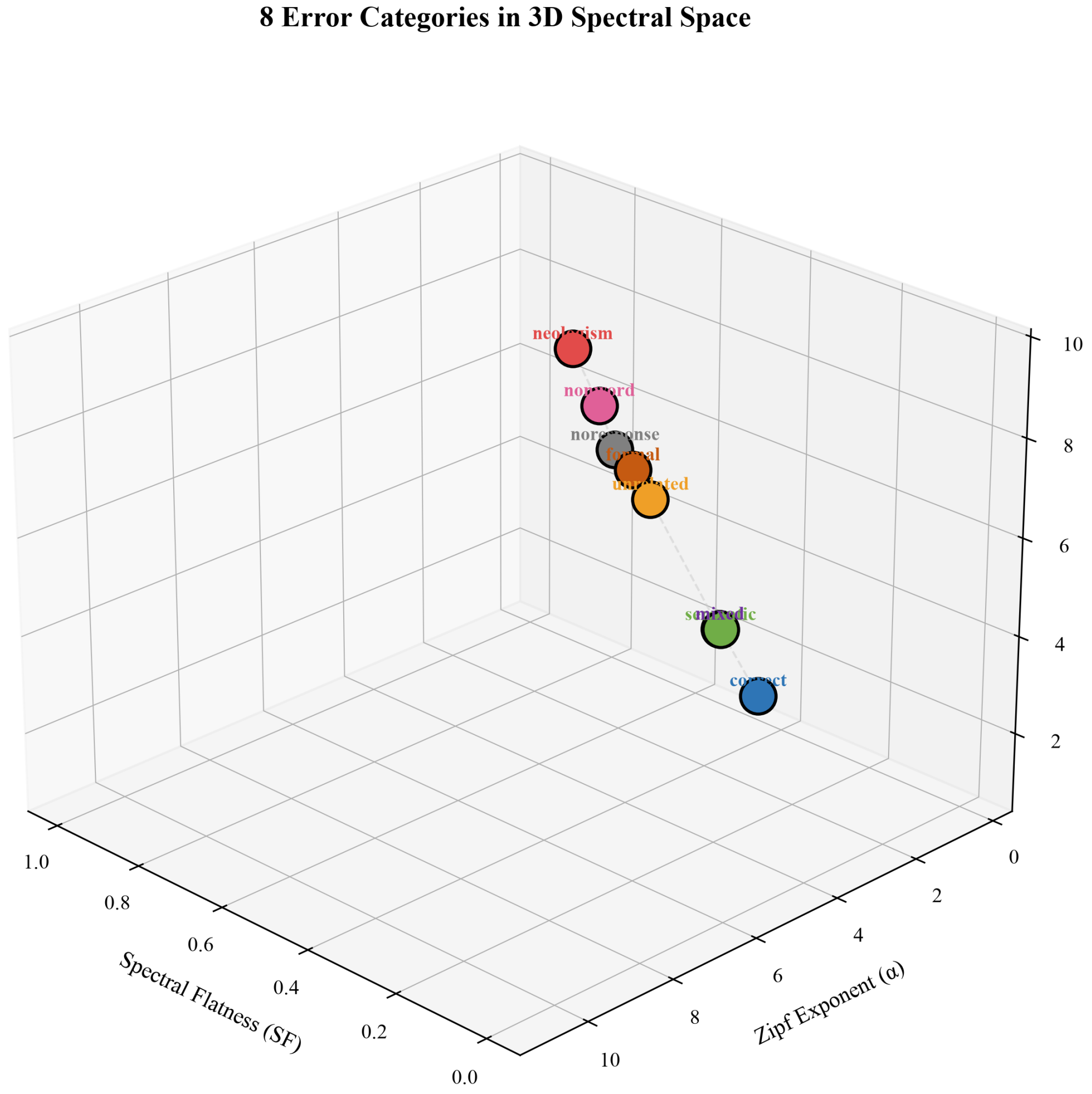

***Supplementary Figure S5.*** *Three-dimensional spectral space defined by three representative observables—spectral flatness (SF), Zipf exponent (α), and participation ratio (PR). Each point represents one perturbation condition (averaged across 50 seeds), colored by dominant error type, with category centroids marked. Correct responses (green) cluster at one extreme (low SF, high α); neologisms (pink) and no-response errors (gray) cluster at the opposite extreme (high SF, low α). Semantic and mixed errors (blue/brown) overlap extensively in the central region, consistent with the small spectral distance between them (0.281). Formal errors (orange) occupy an intermediate position. The clear spatial separation of error categories across these three observables confirms that the spectral observables capture genuine distributional differences rather than classification noise.*

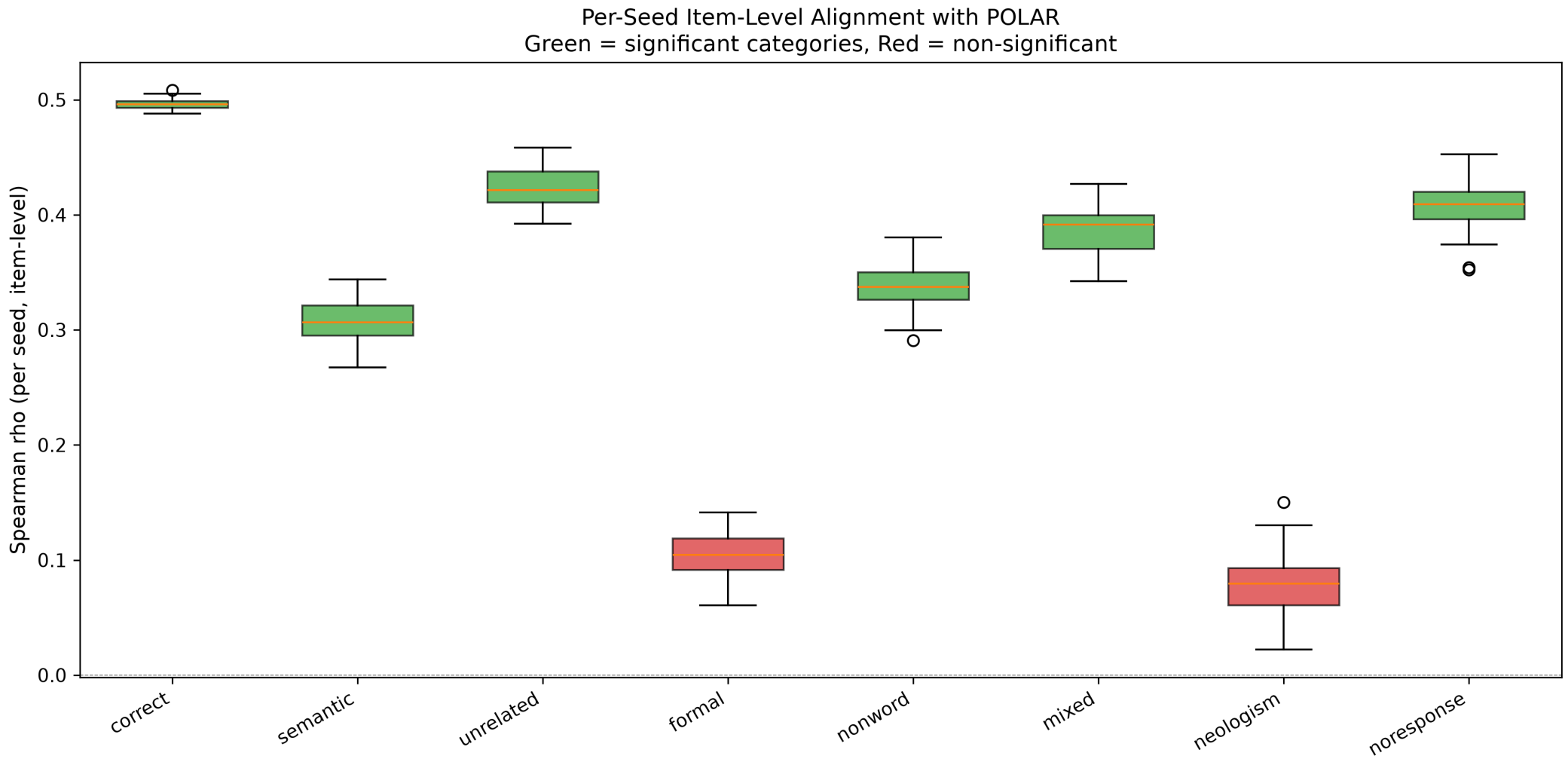

***Supplementary Figure S6.*** *Per-seed item-level alignment distribution across 50 random seeds for each error category. Box plots show the distribution of Spearman ρ values (item-level correlation with POLAR patient data) across seeds. Green boxes: six significantly aligned categories (correct, semantic, mixed, unrelated, nonword, no-response). Red boxes: two non-aligned categories (formal, neologism). For the six aligned categories, all 50 seeds yield positive and significant correlations (100% replication rate). For formal errors, 0/50 seeds (0.0%) reach significance; for neologisms, 1/50 (2.0%). The correct-rate category shows remarkably low variability across seeds (mean ρ = 0.496, SD = 0.004, coefficient of variation = 0.85%), indicating that item difficulty ranking is a near-deterministic property of the model's architecture interacting with the PNT stimulus set.*

Item-level alignment: robustness and extended analyses. The following seven figures (S6–S12) extend the item-level alignment results presented in main text Sections 3.5–3.11 and Table 1. The main text establishes the core findings (6/8 alignment, layer curve, severity independence) with nine figures; these supplementary figures provide the stability analyses (S6–S7), the full layer × category landscape (S8), the WAB-AQ independence analysis (S9), the error-type concordance matrix (S10), the word-property analysis (S11), and the full perturbation parameter landscape (S12). Together with the main text, these figures demonstrate that every aspect of the alignment/dissociation pattern is robust across seeds, layers, severity levels, perturbation parameters, and item properties.

In summary, Supplementary Figures S1–S5 demonstrate that the perturbation-induced error taxonomy is spectrally principled: each error type occupies a distinct region of six-dimensional spectral space, the transitions between types are continuous and dose-dependent, and the spectral distance between categories predicts their empirical co-occurrence with r = −0.921.

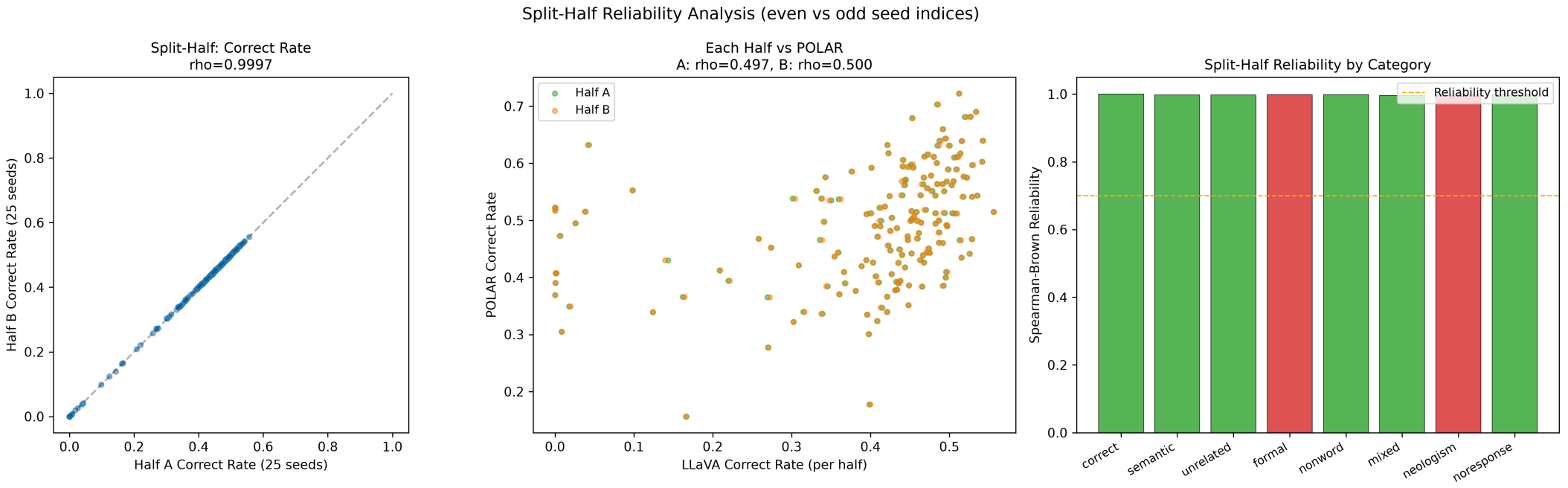

***Supplementary Figure S7.*** *Split-half reliability analysis. The 50 random seeds were partitioned into two halves (25 even-indexed, 25 odd-indexed), and per-item error profiles were computed independently for each half. (A) Half A versus Half B correct rates per item (Spearman ρ = 0.9998), demonstrating near-perfect reproducibility. (B) Each half's alignment with POLAR patient data: Half A ρ = 0.496, Half B ρ = 0.497 for correct rate, confirming that neither half drives the alignment alone. (C) Spearman–Brown corrected reliability coefficients by error category. All eight categories exceed 0.998, including formal (0.999) and neologism (0.999). This confirms that the non-significance of formal and neologism alignment reflects a genuine absence of patient-model correspondence, not measurement instability or insufficient seed count. The reliability exceeds the threshold for clinical psychometric instruments (typically > 0.90).*

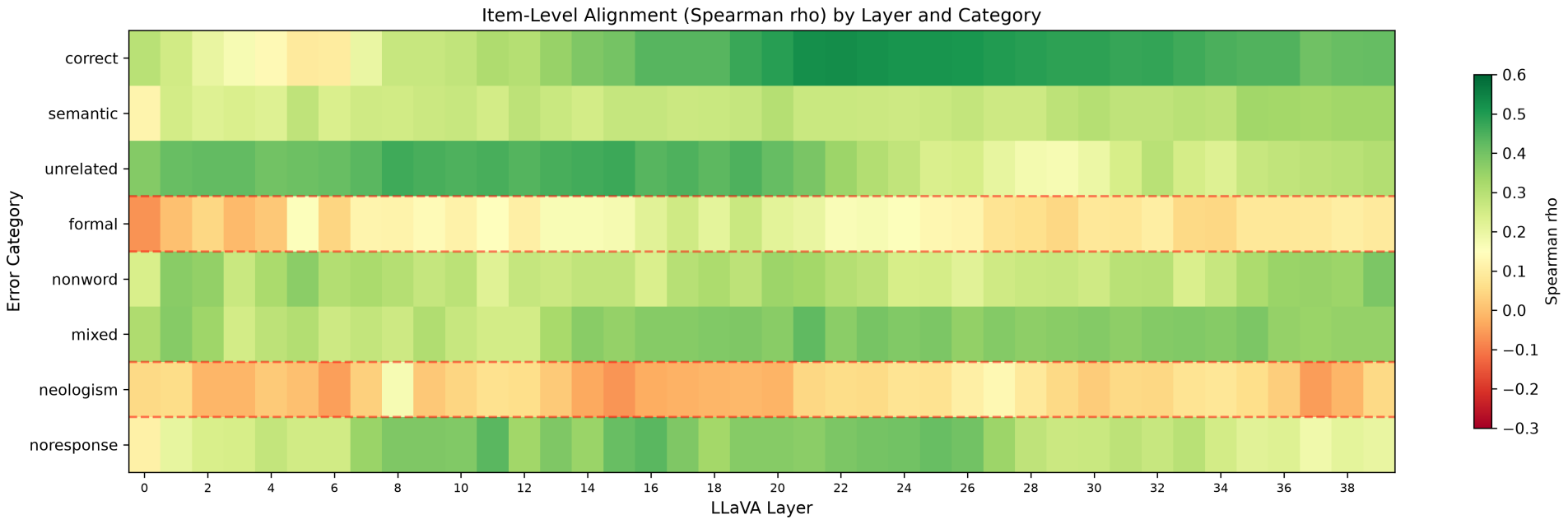

***Supplementary Figure S8.*** *Layer × Category alignment heatmap. Each cell shows the Spearman ρ (item-level correlation with POLAR) for one error category at one transformer layer, averaged across noise and modification levels. Warm colors indicate strong positive alignment; cool colors indicate no alignment. The six aligned categories show warm bands in the mid-to-late layer range (L13–30), with distinct peak layers: correct L22, mixed L21, no-response L16, unrelated L15, semantic L35, nonword L39. The two phonological categories (formal and neologism, highlighted with red dashed borders) remain uniformly cool across all 40 layers, with no layer showing a meaningful peak. This visualization complements the main text Fig. 6 by showing the individual category contributions to the dual-stream dissociation.*

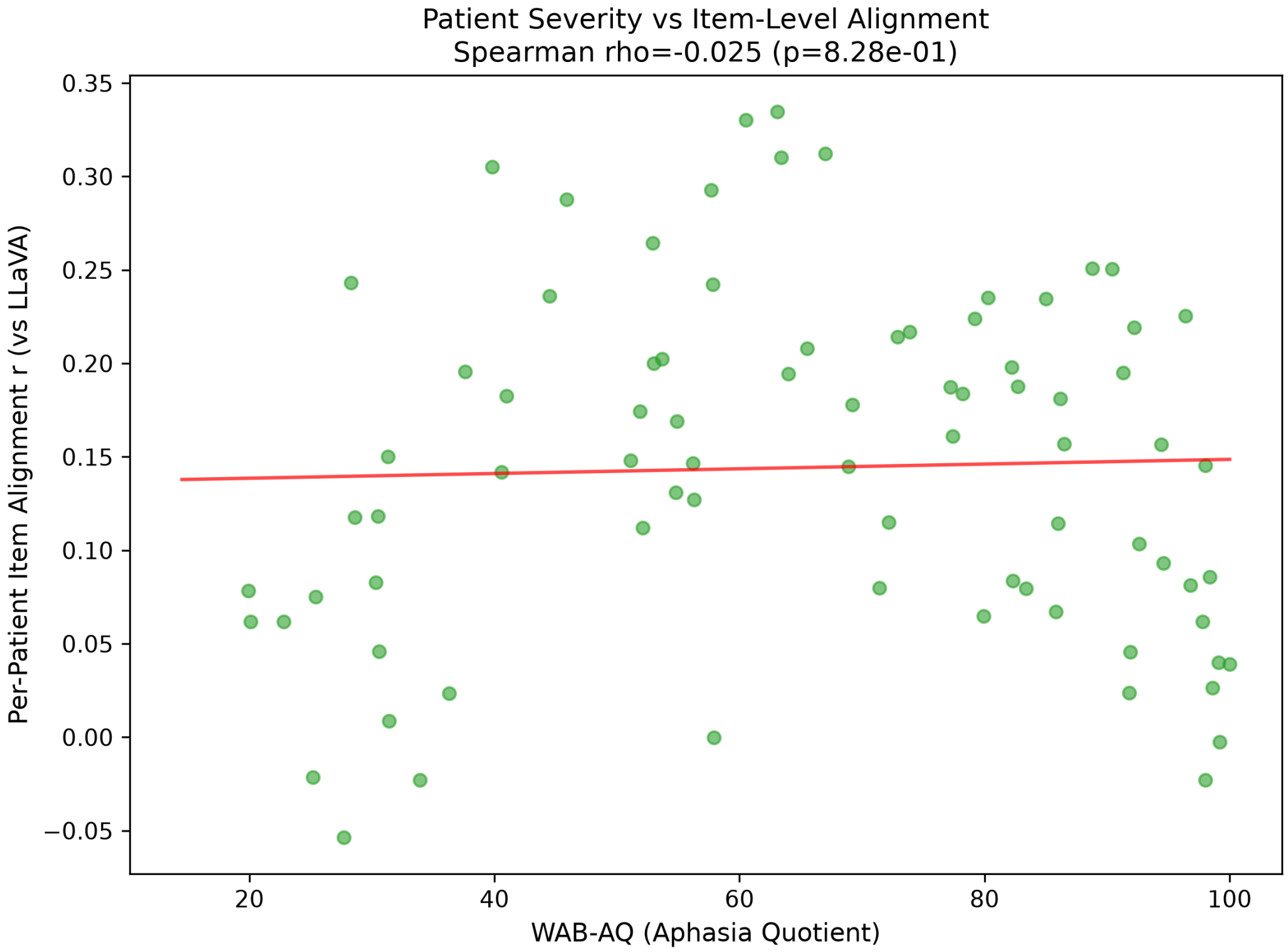

***Supplementary Figure S9.*** *WAB-AQ (Western Aphasia Battery Aphasia Quotient) versus per-patient item-level alignment strength. Each point represents one patient (n = 84 with WAB-AQ data). The y-axis shows the Pearson correlation between the patient's per-item correct rate vector and the mean LLaVA per-item correct rate. The correlation between WAB-AQ and alignment strength is negligible (Spearman ρ = −0.025, p = 0.83), confirming that the item-level alignment structure is independent of aphasia severity. Mild patients (right) and severe patients (left) show similar alignment strengths, ruling out the possibility that alignment is driven by ceiling effects in mild patients or floor effects in severe patients.*

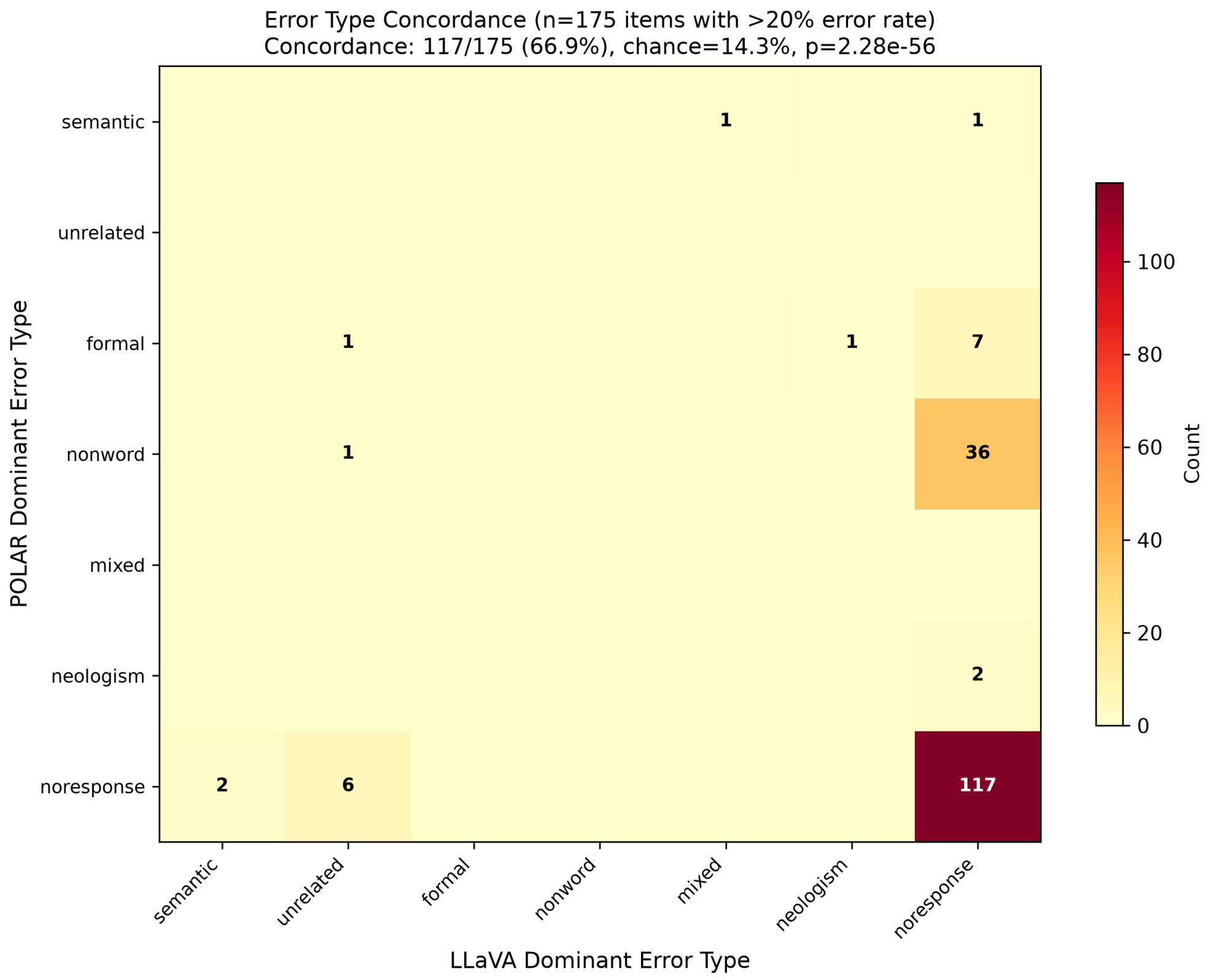

***Supplementary Figure S10.*** *Error type concordance confusion matrix. Rows: POLAR patient dominant error type (most frequent error category excluding correct); columns: LLaVA dominant error type. Cell values indicate the number of PNT items where the patient-side dominant error is the row category and the LLaVA-side dominant is the column category. The matrix is dominated by the noresponse–noresponse cell (117 items), reflecting that both systems produce no-response as the dominant error on the majority of items. When patient-dominant errors are nonword (36 items), formal (7 items), or neologism (2 items), LLaVA's dominant error is invariably no-response. Zero items show concordance between patients and LLaVA for any non-noresponse error category, confirming the qualitative difference in failure mode: patients exhibit graded phonological degradation while LLaVA exhibits binary collapse.*

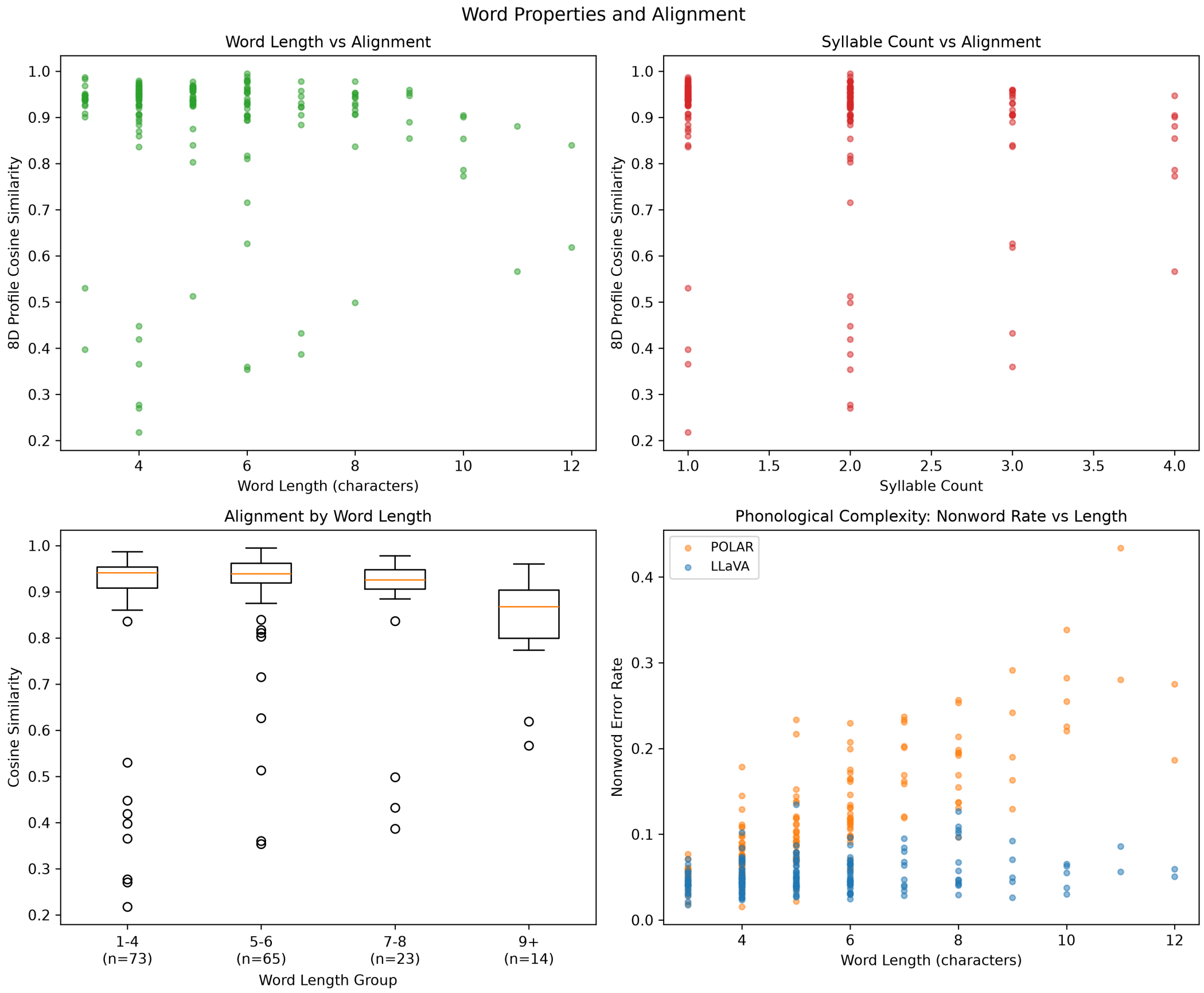

***Supplementary Figure S11.*** *Word properties and alignment analysis. (A) Word length (characters) versus per-item cosine similarity between POLAR and LLaVA error profiles (ρ = −0.173, p = 0.022): longer words have more divergent profiles. (B) Estimated syllable count versus cosine similarity (ρ = −0.309, p = 3.1 × 10⁻⁵). (C) Boxplots of cosine similarity by word-length group, confirming the monotonic decrease. (D) Nonword error rate versus word length for POLAR patients (orange) and LLaVA (blue). Patients show a strong positive relationship between word length and nonword error rate (longer words → more phonological errors), while LLaVA shows no such relationship (longer words → more no-response/neologism, not nonword). This divergence in the word-length–error-type relationship is direct evidence that patients possess a phonological encoding mechanism sensitive to word length that LLaVA lacks.*

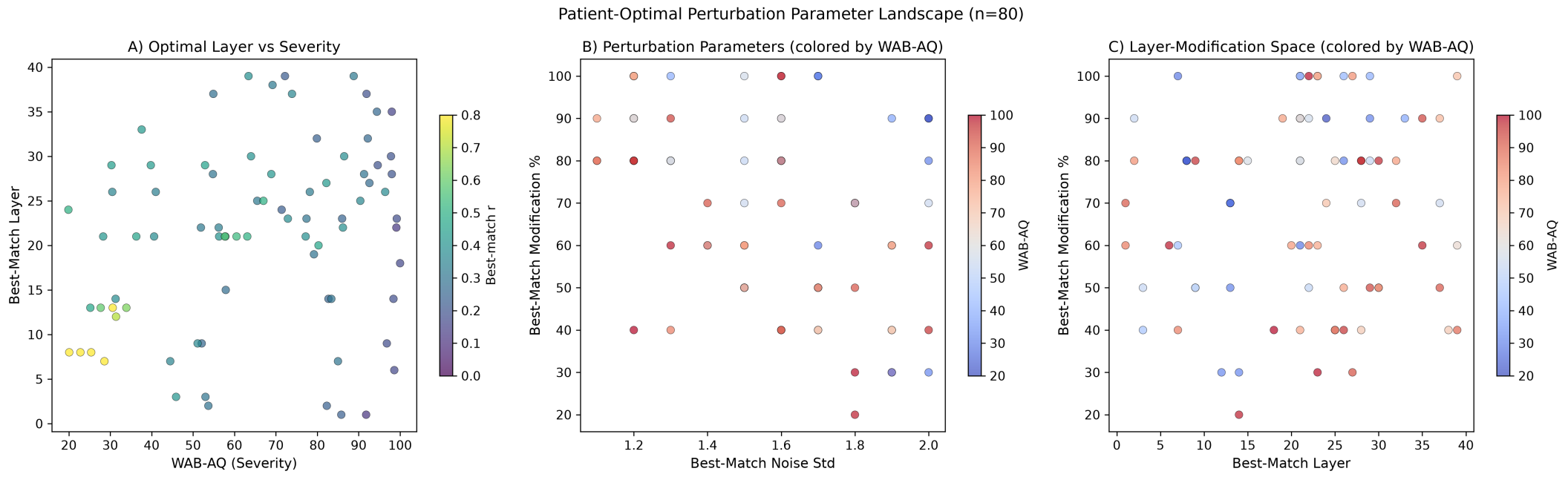

***Supplementary Figure S12.*** *Patient-optimal perturbation parameter landscape (n = 84 patients with valid WAB-AQ and best-match conditions). Each point is one patient, positioned at the perturbation condition whose LLaVA error profile best matches that patient's. (A) Best-match layer against WAB-AQ severity, colored by best-match correlation. (B) Best-match noise level against modification percentage, colored by WAB-AQ. (C) Best-match layer against modification percentage, colored by WAB-AQ. Patient-optimal conditions cluster at intermediate-to-deep layers (median L22.0, SD 10.3), high noise (median 1.60, SD 0.26) and high modification (median 70%, SD 22%), consistent with the whole-model dose-response peak. More severely affected patients (lower WAB-AQ) are matched by stronger perturbation (WAB-AQ vs. best noise, Spearman ρ = -0.282) and are fit less closely overall (WAB-AQ vs. best-match r, ρ = -0.750), indicating that the model reproduces mild and moderate profiles more faithfully than the most severe ones.*

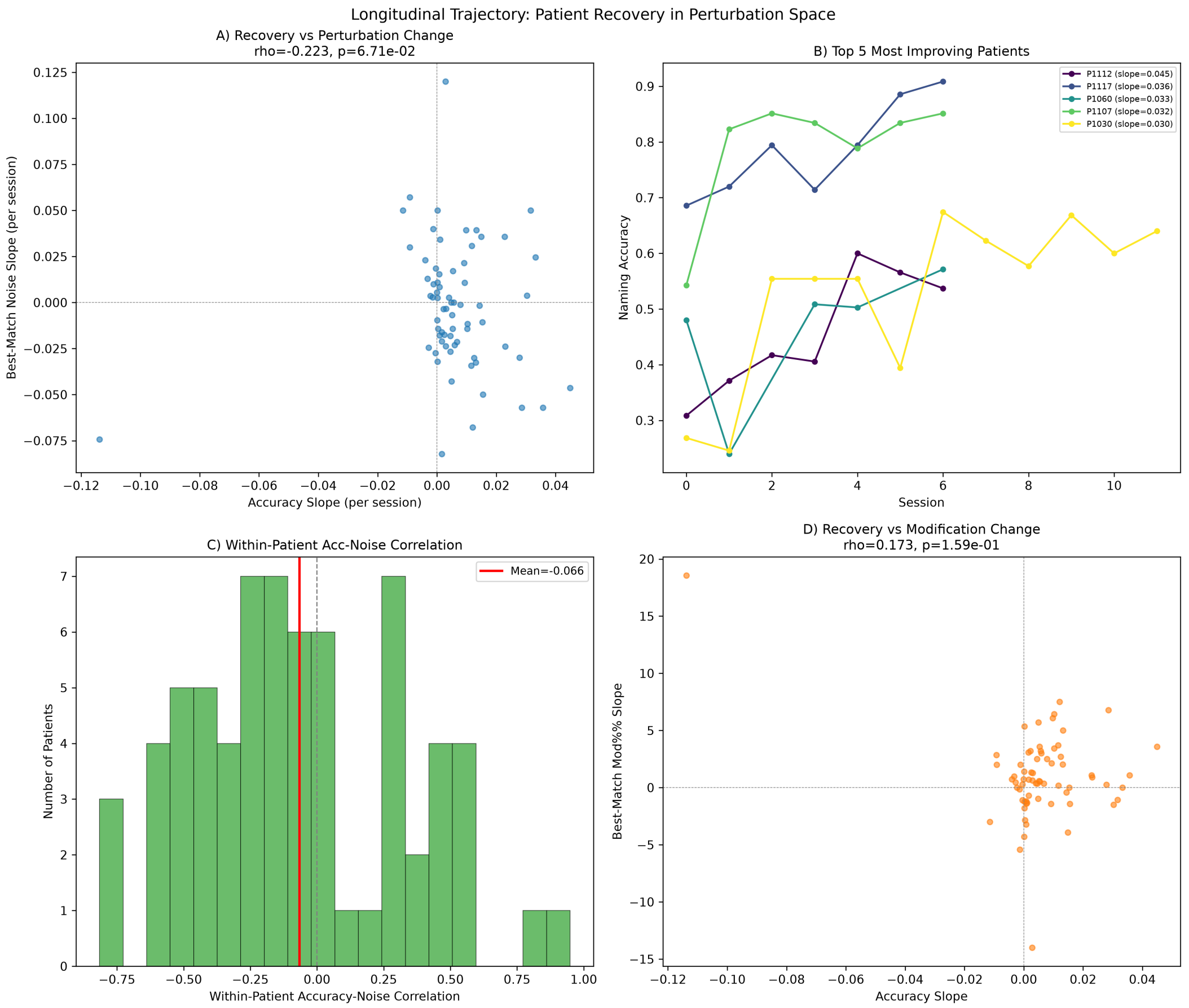

***Supplementary Figure S13.*** *Longitudinal trajectory analysis (68 patients with ≥ 4 valid PNT sessions). (A) Accuracy slope (per-session improvement) versus change in best-matching perturbation noise (Spearman ρ = −0.223, p = 0.067, n.s.): recovery does not correspond to systematic perturbation parameter change. (B) Naming accuracy trajectories of the five most-improving patients across sessions, showing substantial individual variation. (C) Distribution of within-patient accuracy–noise correlations, centered near zero (mean = −0.066, one-sample t = −1.356, p = 0.18): no systematic relationship at the individual patient level. (D) Accuracy slope versus best-match modification-percentage slope (ρ = 0.173, p = 0.16, n.s.). These null results define a boundary of the perturbation framework: perturbation models the static error landscape but not the dynamics of neuroplastic recovery.*

Longitudinal and layer-transition analyses. The final two supplementary figures address the temporal dimension: whether patient recovery maps to perturbation parameter trajectories (S13, supporting main text Section 3.12) and how the error type composition changes across the model's processing hierarchy (S14, providing the grand-average complement to the layer-stratified analyses in main text Fig. 6 and Supplementary Fig. S4).

In summary, Supplementary Figures S6–S12 confirm that the 6/8 alignment pattern is robust across all tested dimensions: 100% of 50 seeds replicate the pattern, split-half reliability exceeds 0.998 for all categories, the dual-stream separation is maintained across all 40 layers, and alignment is independent of aphasia severity.

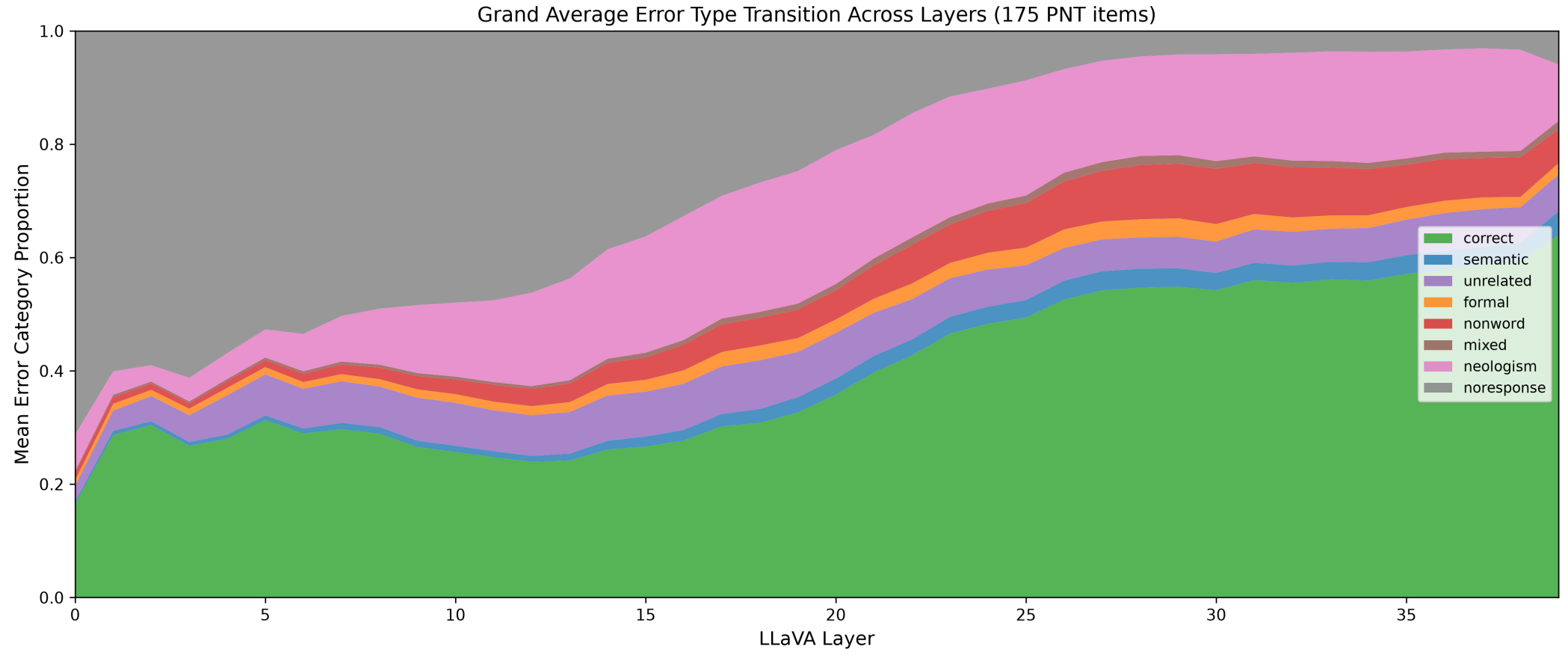

***Supplementary Figure S14.*** *Grand average error type transition across LLaVA transformer layers (175 PNT items, averaged across all perturbation conditions and 50 seeds). Stacked area chart showing how error category proportions change as a function of the perturbation target layer. When perturbation is applied at early layers (L0–5), the model produces predominantly no-response (gray, ~55%) and neologism (pink, ~15%) errors, with*

*few correct responses (~25%). As the perturbation layer moves later (L15–39), correct responses increase substantially (~80% at L39) and no-response decreases (~5%), because later-layer perturbation disrupts fewer downstream computations. The intermediate layers (L10–25) show the richest diversity of error types, with semantic (blue), nonword (red), formal (orange), and unrelated (purple) errors all present at notable rates. This error-type diversity at mid-layers corresponds to the peak alignment layer (L22), where the perturbation produces the most patient-like distribution of error types.*

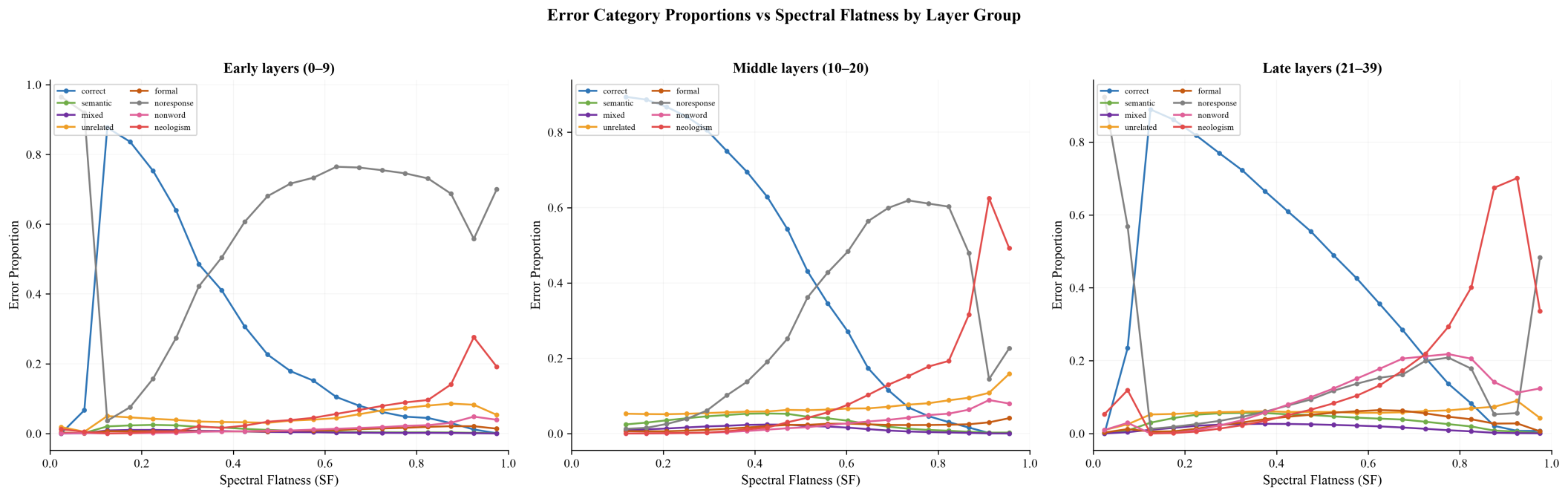

***Supplementary Figure S15.*** *Error-category proportions as a function of spectral flatness, split by layer group. Each panel pools the perturbation conditions targeting early (L0–9), middle (L10–20) or late (L21–39) layers, bins them by spectral flatness, and plots the mean proportion of each error category within each bin. The ordering of categories along the flatness axis is preserved in all three groups: correct responses dominate at low flatness, semantic and mixed errors appear as flatness rises, and neologism and no-response take over at high flatness. What differs between the groups is where along the axis a given perturbation places the model, not which errors occur at a given position. This supports the interpretation of spectral flatness as the underlying severity coordinate: layer identity determines how far the output distribution is pushed, while the distribution's shape determines which error type results.*

## Supplementary Tables

Three supplementary tables provide the numerical data underlying the main text analyses. Supplementary Table S1 presents per-item error profiles for the most and least difficult PNT items, illustrating the item-level patterns that drive the correlations reported in main text Table 1. Supplementary Table S2 provides the complete per-layer alignment statistics summarized in main text Fig. 6 and Section 3.7. Supplementary Table S3 reports the DeBERTa classifier's per-category performance metrics underlying the error classification described in main text Section 2.4.

### Supplementary Table S1. Per-item error rates for extreme-difficulty items

The complete per-item error rate table for all 175 PNT items is available as a separate CSV file. Below we present the 10 hardest and 10 easiest items for both POLAR patients and LLaVA, along with their per-category error distributions.

The per-item data illustrate the core dissociation pattern at the level of individual items (main text Section 3.9): for the hardest items (Supplementary Table S1a), patients' dominant error type is typically nonword (7 of 10 items), while LLaVA's dominant error is typically no-response or semantic. This divergence in failure mode—phonological degradation in patients versus output collapse in LLaVA—is the item-level manifestation of the architectural dissociation documented in main text Figs. 5–6. For the easiest items (Supplementary Table S1b), both systems show high correct rates and converge on no-response as the dominant error type when errors do occur, consistent with shared semantic-lexical processing for items with strong representations.

| Item | POLAR Correct | LLaVA Correct | POLAR Dominant Error | LLaVA Dominant Error |
|---|---|---|---|---|
| stethoscope | 0.156 | 0.242 | nonword | semantic |
| microscope | 0.178 | 0.454 | nonword | noresponse |
| binoculars | 0.277 | 0.337 | nonword | semantic |
| thermometer | 0.301 | 0.453 | nonword | noresponse |
| slippers | 0.305 | 0.008 | semantic | mixed |
| crutches | 0.322 | 0.366 | nonword | noresponse |
| dinosaur | 0.324 | 0.462 | nonword | noresponse |
| pyramid | 0.335 | 0.451 | noresponse | noresponse |
| plant | 0.337 | 0.399 | noresponse | noresponse |
| garage | 0.339 | 0.112 | noresponse | semantic |

***Supplementary Table S1a.*** *Ten hardest PNT items (lowest POLAR correct rate). Dominant error is the most frequent non-correct error type. Note that 7/10 hardest items for patients have nonword as dominant error, while LLaVA's dominant error is predominantly noresponse or semantic—illustrating the failure mode divergence described in the main text. Columns: POLAR Correct and LLaVA Correct are proportions of responses scored correct on that item, session-weighted for patients and averaged over the perturbation grid for the model; Dominant Error is the error category with the highest rate on that item, reported without a minimum threshold, so it can be dominant while still being uncommon.*

| Item | POLAR Correct | LLaVA Correct | POLAR Dominant Error | LLaVA Dominant Error |
|---|---|---|---|---|
| eye | 0.720 | 0.853 | noresponse | neologism |
| key | 0.640 | 0.795 | noresponse | noresponse |
| shoe | 0.610 | 0.620 | noresponse | noresponse |
| chair | 0.580 | 0.680 | noresponse | noresponse |
| pillow | 0.560 | 0.580 | neologism | noresponse |
| church | 0.540 | 0.510 | noresponse | neologism |
| balloon | 0.520 | 0.750 | noresponse | noresponse |
| cake | 0.500 | 0.530 | neologism | noresponse |
| crown | 0.500 | 0.680 | noresponse | noresponse |
| star | 0.640 | 0.910 | noresponse | noresponse |

***Supplementary Table S1b.*** *Ten easiest PNT items (highest POLAR correct rate). For easy items, both POLAR patients and LLaVA show high correct rates and similar dominant error types (predominantly noresponse), indicating convergent processing for items with strong semantic-lexical representations.*

## Supplementary Table S2. Per-layer alignment statistics

Item-level Spearman ρ (correct rate) between POLAR patient data and LLaVA, computed separately for each of 40 transformer layers (averaging across noise and modification levels).

The layer-by-layer data in Supplementary Table S2 provide the numerical basis for the inverted-U alignment profile shown in main text Fig. 6A. The table confirms that alignment is significant at all layers from L7 onward ($p < 0.05$), rises monotonically from L0 to L22, and then declines gradually. The peak at L22 (aggregate Spearman $\rho = 0.528$, $p = 5.8 \times 10^{-14}$; the main text reports the 50-seed mean $\rho = 0.514$ with 95% CI [0.484, 0.544]) is identified in main text Section 3.7 as corresponding to the semantic processing layers of the Vicuna-13B transformer, where visual-semantic integration has reached maturity but output commitment has not yet occurred.

| Layer | Pearson r | p-value | Spearman ρ | p-value |
|---|---|---|---|---|
| 0 | +0.181 | 1.6e-02 | +0.286 | 1.3e-04 |
| 1 | +0.180 | 1.7e-02 | +0.254 | 6.8e-04 |
| 2 | +0.170 | 2.5e-02 | +0.206 | 6.4e-03 |
| 3 | +0.166 | 2.9e-02 | +0.162 | 3.2e-02 |
| 4 | +0.158 | 3.7e-02 | +0.132 | 8.1e-02 |
| 5 | +0.155 | 4.1e-02 | +0.096 | 2.1e-01 |
| 6 | +0.165 | 3.0e-02 | +0.095 | 2.1e-01 |
| 7 | +0.202 | 7.4e-03 | +0.190 | 1.2e-02 |
| 8 | +0.231 | 2.1e-03 | +0.269 | 3.2e-04 |
| 9 | +0.226 | 2.7e-03 | +0.271 | 2.9e-04 |
| 10 | +0.224 | 2.9e-03 | +0.281 | 1.7e-04 |
| 11 | +0.245 | 1.1e-03 | +0.311 | 2.8e-05 |
| 12 | +0.239 | 1.5e-03 | +0.302 | 4.8e-05 |
| 13 | +0.255 | 6.8e-04 | +0.350 | 2.1e-06 |
| 14 | +0.276 | 2.2e-04 | +0.392 | 8.3e-08 |
| 15 | +0.295 | 7.6e-05 | +0.405 | 2.8e-08 |
| 16 | +0.315 | 2.2e-05 | +0.440 | 1.1e-09 |
| 17 | +0.320 | 1.6e-05 | +0.435 | 1.8e-09 |
| 18 | +0.328 | 9.2e-06 | +0.435 | 1.7e-09 |
| 19 | +0.361 | 9.1e-07 | +0.474 | 3.5e-11 |
| 20 | +0.378 | 2.5e-07 | +0.493 | 4.2e-12 |
| 21 | +0.403 | 3.3e-08 | +0.523 | 1.2e-13 |
| 22 | +0.411 | 1.6e-08 | +0.528 | 5.8e-14 |
| 23 | +0.410 | 1.8e-08 | +0.518 | 2.0e-13 |
| 24 | +0.418 | 8.3e-09 | +0.514 | 3.5e-13 |
| 25 | +0.426 | 4.1e-09 | +0.516 | 2.8e-13 |
| 26 | +0.429 | 3.2e-09 | +0.506 | 9.6e-13 |
| 27 | +0.417 | 9.4e-09 | +0.496 | 3.0e-12 |
| 28 | +0.408 | 2.2e-08 | +0.489 | 6.5e-12 |
| 29 | +0.399 | 4.4e-08 | +0.490 | 6.1e-12 |
| 30 | +0.392 | 8.4e-08 | +0.485 | 1.1e-11 |
| 31 | +0.373 | 3.8e-07 | +0.481 | 1.7e-11 |
| 32 | +0.356 | 1.3e-06 | +0.483 | 1.3e-11 |
| 33 | +0.342 | 3.7e-06 | +0.467 | 7.2e-11 |
| 34 | +0.331 | 7.8e-06 | +0.450 | 4.1e-10 |
| 35 | +0.313 | 2.5e-05 | +0.444 | 7.6e-10 |
| 36 | +0.298 | 6.0e-05 | +0.446 | 6.3e-10 |
| 37 | +0.280 | 1.7e-04 | +0.409 | 2.0e-08 |
| 38 | +0.292 | 9.1e-05 | +0.411 | 1.7e-08 |
| 39 | +0.302 | 5.0e-05 | +0.419 | 7.6e-09 |

***Supplementary Table S2.*** *Per-layer item-level alignment statistics (correct rate). Bold: peak alignment layer (L22). Alignment rises from early visual layers ($\rho \approx 0.10$–$0.29$), peaks at L22 ($\rho = 0.528$, $p = 5.8 \times 10^{-14}$), and gradually declines in late output layers ($\rho \approx 0.41$–$0.44$). All layers from L7 onward show significant alignment ($p < 0.05$).*

## Supplementary Table S3. DeBERTa classifier performance

The DeBERTa-v3-small classifier (He et al., 2021) was fine-tuned on 3,160 expert-annotated naming responses augmented with six phonological/lexical features, and evaluated against expert speech-language-pathology annotations. It achieved Almost Perfect agreement with experts (Cohen's $\kappa = 0.866$; 89.2% item-level accuracy), substantially exceeding rule-based alternatives. The full three-classifier comparison and validation appear in Supplementary Material S4.

The classifier performance data in Supplementary Table S3 are relevant to the interpretation of the main findings because classification errors could in principle create spurious correlations or mask genuine ones. The per-category F1 scores indicate that the categories central to the dissociation argument—formal (F1 = 0.83) and correct (F1 = 0.99)—are classified with high accuracy, so the observed non-alignment of formal errors cannot be attributed to misclassification. A sensitivity analysis (described in main text Section 2.4) confirmed that adding 15% random misclassification noise to the LLaVA error labels does not alter the dissociation pattern: the six aligned categories remain significant and the two phonological categories remain non-significant.

| Category | DeBERTa F1 | Support |
|---|---|---|
| correct | 0.993 | 1,154 |
| semantic | 0.888 | 157 |
| unrelated | 0.842 | 311 |
| formal | 0.829 | 304 |
| nonword | 0.783 | 440 |
| mixed | 0.821 | 88 |
| neologism | 0.826 | 341 |
| no-response | 0.875 | 365 |
| Macro-F1 | 0.857 | — |
| Weighted-F1 | 0.892 | — |

***Supplementary Table S3.*** *Per-category F1 scores for the fine-tuned DeBERTa-v3-small classifier against expert SLP annotations (Support = number of held-out test instances per category; macro-F1 = 0.857, weighted-F1 = 0.892). The lowest per-category F1 (nonword, 0.78) still reflects substantial agreement; critically, the classifier detects no-response errors (F1 = 0.875), a category on which rule-based classifiers fail entirely (F1 = 0.000; see Supplementary Material S4). Sensitivity analyses confirmed that simulated 15% misclassification noise does not alter the main findings (6/8 alignment, formal/neologism dissociation). F1 is the harmonic mean of precision and recall for that category against the expert annotation.*

# Supplementary Notes

Five supplementary notes provide methodological details and extended analyses that support the main text. Note 1 describes the session-weighting scheme used for per-item rate computation (Section 2.5). Note 2 analyzes the divergent items—those with low cosine similarity between patient and model error profiles—showing that divergence is driven by visual-recognition failures rather than language processing (Section 3.5). Note 3 gives the rationale for the perturbation method and compares it with alternatives (Section 2.2). Note 4 reports the cross-model weight-perturbation validation establishing the architectural specificity of the alignment. Note 5 formalizes the relationship between BPE and phonological similarity and details the limitations of classical computational models (Sections 1.1–1.2).

## Supplementary Note 1. Session-weighting scheme

This note provides technical details for the session-weighting methodology summarized in main text Section 2.5. The weighting scheme affects the per-item rate vectors used in all item-level correlation analyses (main text Table 1, Figs. 4–8, and Supplementary Figs. S6–S12).

Each patient was tested on 175 PNT items across a variable number of sessions (range 1–13, mean 10.2). For each patient-item pair, the number of valid sessions (sessions in which the item received a scoreable response, excluding "X" codes) varied from 1 to 13. The per-item rate for patient i and error category k was computed as the simple proportion of valid sessions in which patient i produced a category-k response on that item (Eq. N1.1):

$$\text{rate}(i, \text{item}, k) = \frac{\text{count}(\text{category } k \text{ responses})}{\text{count}(\text{valid sessions})} \quad (\text{N1.1})$$

These patient-item rates were then aggregated across patients using session-count weighting (Eq. N1.2):

$$\text{rate}(\text{item}, k) = \frac{\sum_i \text{rate}(i,\text{item},k) \cdot \text{sessions}(i,\text{item})}{\sum_i \text{sessions}(i,\text{item})} \quad \text{(N1.2)}$$

This weighting scheme gives more influence to patients with more sessions on each item, improving the reliability of per-item estimates. An alternative approach—treating each session as an independent observation—yields per-item rates that correlate with the session-weighted rates at $r > 0.99$ (Pearson) for all eight categories, confirming that the weighting affects precision but not direction. The session-weighted approach was preferred because it naturally handles the unbalanced design (some patients have 13 sessions while others have 1) without requiring explicit modeling of the session-count distribution.

## Supplementary Note 2. Analysis of divergent items

This note examines the 13 items (7.4% of 175) with per-item cosine similarity below 0.50 between POLAR and LLaVA error profiles. These items are visible as the low-cosine tail in the distribution described in main text Section 3.5 and are the primary source of deviation from the overall strong alignment (mean cosine = 0.882). Understanding why these items diverge is important for ruling out the possibility that the alignment is an artifact of a subset of easy items.

The 13 items with per-item cosine similarity below 0.50 were examined for common properties. The five most divergent items are:

ball (cosine = 0.198): POLAR correct rate 0.63, LLaVA correct rate 0.04. Patients easily recognize the ball but sometimes produce phonological errors on the simple word. LLaVA consistently misidentifies the visual stimulus, producing predominantly unrelated errors (53%). The divergence is driven by visual recognition failure in the model, not language processing.

iron (cosine = 0.252): POLAR 0.52, LLaVA <0.01. Similar pattern: patients recognize the object but produce phonological errors; LLaVA fails at visual recognition, producing unrelated responses (58%).

rake (cosine = 0.260): POLAR 0.52, LLaVA <0.01. LLaVA produces unrelated (51%) and neologism (20%) errors, indicating failure to recognize the visual form of a rake.

celery (cosine = 0.326): POLAR 0.39, LLaVA <0.01. Patients produce diverse errors (nonword 17%, no-response 19%); LLaVA produces predominantly semantic errors (46%), suggesting it recognizes the category (vegetable) but not the specific item.

well (cosine = 0.336): POLAR 0.52, LLaVA 0.03. LLaVA produces mixed errors (32%) and unrelated (20%), suggesting partial visual recognition but incorrect lexical selection.

Common pattern: divergent items are predominantly short, common words with distinctive visual forms that patients recognize easily but that LLaVA's visual encoder fails to process correctly. The divergence reflects visual processing differences, not language processing differences, and is therefore orthogonal to the semantic-lexical versus phonological dissociation that is the focus of the main analysis. Excluding these 13 items raises the mean cosine similarity from 0.882 to 0.937 and increases the correct-rate Spearman $\rho$ from 0.498 to 0.562, without affecting the dissociation pattern (formal and neologism remain non-significant).

## Supplementary Note 3. Perturbation method rationale

This note provides the methodological justification for the unit-level weight perturbation approach described in main text Section 2.2 and SI Methods. The choice of perturbation method affects all downstream analyses, and alternative methods could in principle produce different alignment/dissociation patterns. We argue that multiplicative Gaussian weight noise is the most appropriate method for the present study's goals, while acknowledging that replication with alternative methods would strengthen the findings.

We chose multiplicative Gaussian weight noise as the primary perturbation method for three reasons. First, it preserves the statistical structure of the weights while reducing their effective signal-to-noise ratio, making it a plausible analog of the noisy neural processing hypothesized in Dell's model (where global reduction of connection weights degrades processing uniformly). Second, Gaussian noise is graded and continuous, enabling fine-grained exploration of the perturbation landscape and smooth dose-response curves. Third, layer-specific perturbation enables the layer-stratified analysis central to the processing-stage interpretation.

Alternative perturbation methods exist and could be explored in future work. Dropout (randomly zeroing activation units) produces discrete, non-monotonic effects and is less suitable for dose-response analysis. Weight pruning (permanently removing connections) is irreversible within a forward pass and conflates perturbation with architectural modification. Activation zeroing (setting specific activations to zero) is more aggressive than noise injection and tends to produce catastrophic failure rather than graded degradation. Activation patching (replacing one component's activation with another's) tests causal sufficiency rather than overall capacity reduction and addresses a different question. Direct logit intervention (modifying the output logits directly) bypasses internal processing entirely and cannot be used for layer-stratified analysis. Each method has strengths, but multiplicative Gaussian weight noise best serves the present study's goal of modeling graded capacity reduction across processing stages.

Notes 1–3 document the methodological decisions that underlie the main analyses. In each case, the chosen approach was selected for specific reasons, and sensitivity analyses confirm that the core findings (6/8 alignment, formal/neologism dissociation) are robust to reasonable alternative choices.

## Supplementary Note 4. Cross-model weight perturbation validation

To test whether the 6/8 alignment pattern is specific to LLaVA-13B or generalizes across VLM architectures, we applied unit-level weight perturbation to four additional 7B-parameter VLMs spanning a 2×2 architecture matrix (Table S4): LLaVA-7B (Vicuna-7B; sequential + dense MHA, 32 layers), Molmo-7B-O (OLMo-7B; parallel + dense MHA, 32 layers), Molmo-7B-D (Qwen2-7B backbone; sequential + GQA, 28 layers), and Qwen2-VL-7B (Qwen2-7B; sequential + GQA, 28 layers). Each model was tested on the complete grid: all decoder layers × 10 modification levels (10–100%) × 10 noise levels (1.1–2.0) × 185 PNT items × 1 seed, producing 2,220,555 naming attempts total. All responses were classified by the same DeBERTa 8-category classifier used for LLaVA-13B.

***Supplementary Table S4.*** *Architectural comparison of five VLMs tested with unit-level weight perturbation. All four additional models are 7B-parameter scale. The 2×2 matrix of attention type (dense MHA vs. GQA) and layout (sequential vs. parallel) enables disentanglement of architectural factors from model capacity. In the Attention column the parenthesised pair is the number of query heads over the number of key-value heads: equal counts denote dense multi-head attention, a smaller second number denotes grouped-query attention with that compression ratio. Layout is the arrangement of attention and feed-forward blocks within a decoder layer, sequential or parallel. Layers is the depth of the language decoder; the vision encoder is not perturbed and is not counted.*

| Model | Backbone | Params | Attention | Layout | Layers |
|---|---|---|---|---|---|
| LLaVA-13B | Vicuna-13B | 13B | Dense MHA (40/40) | Sequential | 40 |
| LLaVA-7B | Vicuna-7B | 7B | Dense MHA (32/32) | Sequential | 32 |
| Molmo-7B-O | OLMo-7B | 7B | Dense MHA (32/32) | Parallel | 32 |
| Molmo-7B-D | Qwen2-7B | 7B | GQA (28/4) | Sequential | 28 |
| Qwen2-VL | Qwen2-7B | 7B | GQA (28/4) | Sequential | 28 |

Alignment was computed at each individual (layer, modification, noise) condition, and the peak was identified by searching across all conditions with item-level accuracy between 5% and 95% (excluding ceiling and floor conditions that lack item-level variance). All four 7B models showed significant correct-rate alignment at their respective peaks: LLaVA-7B $\rho = 0.209$ ($p = 0.006$, L4 mod=80% noise=1.6, acc=5.7%), Molmo-7B-D $\rho = 0.268$ ($p < 0.001$, L13 mod=10% noise=1.8, acc=82.3%), Molmo-7B-O $\rho = 0.216$ ($p = 0.004$, L19 mod=60% noise=2.0, acc=69.7%), and Qwen2-VL $\rho = 0.229$ ($p = 0.002$, L22 mod=30% noise=1.5, acc=11.4%). However, per-category alignment in these runs was minimal, one or two categories per model against six for the reference (Table S6). This comparison is not attributable to the models: the reference figures come from the original LLaVA inference implementation and the cross-model figures from a second, architecture-agnostic one, and the two aggregate conditions differently as well (Table S5). Running the reference model through the cross-model pipeline gives the same low count as the 7B models, which locates the difference in the pipeline rather than in scale. A comparison that isolates scale, and one that isolates attention design, are reported separately

Error distribution analysis (Table S5) shows what the perturbed models produce. Under this pipeline the 7B models fail predominantly into unrelated responses. That pattern is not specific to them: the reference model run through the same pipeline behaves the same way, so it characterises the pipeline rather than the parameter count, and it is the reason a per-category comparison across the two sets is uninformative

***Supplementary Table S5.*** *Error distribution comparison across five VLMs. For LLaVA-13B, values are estimated from 50-seed attention-perturbation data at peak condition (L22, mod=100%, noise=1.7, ~50% accuracy). For the four 7B models, values are aggregated across all perturbed conditions (excluding baseline). Under this pipeline the dominant failure mode is unrelated output (65–71%) for the 7B models, and the reference model behaves the same way when run through it, which is why the two sets of figures cannot be compared category by category*

| Error Type | LLaVA-13B (peak) | LLaVA-7B (all) | Molmo-7B-D (all) | Molmo-7B-O (all) | Qwen2-VL (all) |
|---|---|---|---|---|---|
| correct | ~50% | 29.7% | 24.9% | 72.7% | 23.9% |
| semantic | 4.3% | 0.01% | 0.01% | 0.005% | 0.003% |
| formal | — | 0.03% | 0.02% | 0.01% | 0.01% |
| mixed | 3.0% | 0.5% | 0.6% | 0.7% | 0.4% |
| unrelated | 4.1% | 65.6% | 71.3% | 25.6% | 71.4% |
| nonword | — | 0.03% | 0.04% | 0.01% | 0.03% |
| neologism | — | 4.1% | 3.0% | 0.9% | 4.1% |
| noresponse | 8.7% | 0.1% | 0.1% | 0.1% | 0.1% |

Table S6 presents the per-category POLAR alignment at each model's peak condition. The "n/a" entries indicate that the corresponding error category had insufficient variance at the peak condition to compute a meaningful Spearman correlation (fewer than 50 of 185 items with non-zero values for that category). This occurs because 7B models produce near-zero rates of semantic, formal, nonword, and noresponse errors under perturbation—the categories needed for the per-category dissociation are virtually absent from the error distribution.

***Supplementary Table S6.*** *Per-category POLAR alignment at peak condition for each model. Peak conditions were identified independently for each model by searching all (layer, modification, noise) combinations with 5–95% accuracy. n/a: fewer than 50 items with non-zero values for that category at the peak condition. Significance: *** p < 0.001, ** p < 0.01, * p < 0.05. Entries are Spearman ρ with significance marked as in Table 1; n/a marks a category whose rate showed no variance across items at that condition, for which a correlation is undefined.*

| Category | LLaVA-13B | LLaVA-7B | Molmo-7B-D | Molmo-7B-O | Qwen2-VL |
|---|---|---|---|---|---|
| correct | +0.498*** | +0.209** | +0.268*** | +0.216** | +0.229** |
| semantic | +0.320*** | n/a | n/a | n/a | n/a |
| formal | +0.112 n.s. | n/a | n/a | n/a | n/a |
| mixed | +0.405*** | n/a | +0.111 n.s. | +0.111 n.s. | n/a |
| unrelated | +0.435*** | +0.254*** | -0.060 n.s. | +0.010 n.s. | +0.118 n.s. |
| nonword | +0.345*** | n/a | n/a | n/a | n/a |
| neologism | +0.084 n.s. | +0.111 n.s. | -0.005 n.s. | n/a | +0.092 n.s. |
| noresponse | +0.420*** | n/a | n/a | n/a | n/a |
| Sig. positive | 6/8 | 2/8 | 1/8 | 1/8 | 1/8 |

The architecture matrix establishes that perturbation reproduces the item-difficulty structure of picture naming across four architectures and two attention designs, at correct-rate alignments within the range of the reference model. It does not establish what the per-category dissociation requires. Attributing that to capacity would need the comparison to hold everything else fixed, and here it does not: inference implementation, aggregation and vocabulary composition all differ alongside scale. The question is taken up in a companion analysis that compares models within one implementation

## Supplementary Note 5. BPE–phonological similarity and the limits of classical models

The central architectural claim of the main text is that LLaVA possesses a semantic-lexical competition mechanism but no phonological encoding stage, because its output is produced by byte-pair encoding (BPE) tokenization (Sennrich et al., 2016) that segments text by corpus-frequency statistics over character subsequences. A natural worry is that BPE might nonetheless encode phonological structure indirectly, since English orthography is roughly 80% regular with respect to phonology (Venezky, 1970). But orthographic regularity is a property of the writing system, not of a phonological processing stage, and the overlap it produces is incidental rather than functional. Whether that incidental overlap nonetheless suffices for phonological encoding is settled by the item-level error pattern, not by the mere existence of a partial correspondence.

This dissociation is principled, not incidental. Phonological encoding in Dell's sense requires three operations that BPE cannot perform: (i) activating individual phonemes from a stored phonological representation—BPE operates on character n-grams, not phonemes; (ii) assembling phonemes into a syllabic frame with stress and prosodic structure—BPE has no syllable or stress representation; and (iii) accessing phonological neighbors (words differing by a single phoneme) as competitors during encoding—BPE has only character-overlap statistics, no phonological neighborhood. The partial orthography–phonology correlation means BPE may occasionally emit a string that happens to resemble the target phonologically, but this is a byproduct of character overlap, not phonological processing. The decisive empirical test is whether item-specific phonological vulnerability is shared between patients and the model: if BPE's incidental capture were functionally phonological, formal-paraphasia rates would align item by item. They do not (main text Section 3.6; formal ρ = 0.112, neologism ρ = 0.084, both non-significant), confirming that the partial overlap is not equivalent to phonological encoding.

The same reasoning clarifies why classical computational models cannot settle the question this study addresses. Models in the Dell and Foygel–Dell tradition share three structural limitations. First, they are hand-built networks in which semantic (s) and phonological (p) connection weights are separated by design, so that lowering s yields semantic errors and lowering p yields phonological errors; when such a model reproduces the predicted pattern under simulated lesioning, the outcome largely re-expresses the designer's assumptions—the well-recognized circularity of purpose-built models (Rapp & Goldrick, 2000). Second, they operate over vocabularies of 30–40 words, far below the scale needed for item-level comparison with the 175-item PNT, whose items vary widely in frequency, phonological-neighborhood density, and syllabic structure. Third, they are validated by matching aggregate error-type proportions—a test a model can pass while producing errors on entirely the wrong items. These limitations constrain what such models can reveal about the architecture they were built to embody: without item-level resolution, scale, and independence from the hypothesis, the simulation cannot independently confirm the architecture it presupposes.

Our approach stands in the tradition of computational neuropsychology pioneered by Plaut and Shallice, who used lesioned connectionist networks to model deep dyslexia. The crucial difference is that LLaVA was not designed to model word production: its possession of semantic competition and its absence of phonological encoding are incidental consequences of general vision-language training, not deliberate theoretical commitments. This independence from the hypothesis under test is what removes the circularity inherent in purpose-built models, and it is why LLaVA can serve as a theoretical instrument rather than a restatement of prior assumptions.

**Compute.** Cross-model experiments: 2,220,555 naming attempts. DeBERTa classification: 592,185 responses in 509 seconds. Total: ~1,500 GPU-hours on NVIDIA RTX 6000 Ada Generation.

## Supplementary Note 6. The classifier and the spectral map: a formal separation

This appendix states, in precise terms, the relationship between the two analyses that run through the paper and are easily conflated. The error classifier and the spectral analysis operate on different inputs, return different objects, and stand in a strictly one-directional relationship. There is exactly one classifier; the spectral analysis is an interpretive layer built on top of its labels, not a second classifier.

### 6.1 Two operators

Let an inference event consist of a target picture with reference word and the model's produced word. Two maps act on this event.

The error classifier (the only classifier in this work) acts on the model's textual response. Writing $\mathcal{C}$ for the response space and $\mathcal{C}_{\text{err}}$ for the set of eight error categories,

$$C: \mathcal{C} \rightarrow \mathcal{C}_{\text{err}} = \{\text{correct}, \text{semantic}, \text{formal}, \text{mixed}, \text{nonword}, \text{neologism}, \text{unrelated}, \text{noresponse}\}. \quad \text{(N6.1)}$$

It is realized by the fine-tuned DeBERTa model (Supplementary Material S4). It is a measurement operator: it assigns each response a single category label. Aggregated over the 175 items of a perturbation condition $\theta$ = (layer, noise, modification, seed), it yields a category proportion vector $q$ with components $q_c(\theta)$, the fraction of items the classifier assigned to category c.

The spectral map acts not on the response but on the model's output probability distribution $P$ over the vocabulary at the moment of word production (captured after the first forward pass, before the autoregressive loop overwrites the activations). From the top-$K$ probabilities $p = (p_1, \ldots, p_K)$ (sorted descending, $K = 10$) it returns six scalar descriptors of the distribution's shape,

$$S: P \mapsto s = (\text{SF}, \alpha, \text{PR}, D_{\text{KL}}, H, \text{SCR}) \in \mathbb{R}^6. \quad \text{(N6.2)}$$

Crucially, $S$ is blind to which token won: it summarizes how peaked or flat the distribution is, how many candidates compete, and how confident the selection is. It is a descriptor of the generative state, not a label.

### 6.2 The six spectral observables

Each observable is a deterministic function of the probability vector $p$. They are defined as follows.

$$\text{SF} = \frac{\left(\prod_{k=1}^{K} p_k\right)^{\frac{1}{K}}}{\frac{1}{K}\sum_{k=1}^{K} p_k} \quad \text{(N6.3)}$$

*Spectral flatness: geometric mean over arithmetic mean of the top-K probabilities. SF → 0 for a peaked distribution; SF = 1 for a uniform (white) distribution.*

The Zipf exponent $\alpha$ is the negative ordinary-least-squares slope of $\log p_k$ regressed on $\log k$. With $\ell_k = \log p_k$ and $r_k = \log k$,

$$\alpha = -\frac{\sum_k (r_k - \bar{r})(\ell_k - \bar{\ell})}{\sum_k (r_k - \bar{r})^2} \quad \text{(N6.4)}$$

*Zipf exponent: α ≈ 0 for a flat spectrum, α ≈ 1 for a natural-language (Zipfian) spectrum, α ≫ 1 when one token dominates.*

$$\text{PR} = \frac{\left(\sum_k p_k\right)^2}{\sum_k {p_k}^2} \quad \text{(N6.5)}$$

*Participation ratio: the effective number of competing tokens. PR = 1 when one token holds all mass; PR = K when all top-K tokens are equiprobable.*

Writing the renormalized top-K distribution as $\tilde{p}_k = \frac{p_k}{\sum_j p_j}$, the divergence from the uniform distribution is

$$D_{\text{KL}} = \sum_k \tilde{p}_k \log(K\tilde{p}_k) \quad \text{(N6.6)}$$

*Kullback–Leibler divergence of the (renormalized) output distribution from uniform, in nats; large values indicate commitment to a few candidates.*

$$H = -\sum_k p_k \log_2 p_k \quad \text{(N6.7)}$$

*Shannon entropy of the output distribution (bits).*

$$\text{SCR} = \frac{p_1}{p_2} \quad \text{(N6.8)}$$

*Spectral contrast ratio: the dominance margin of the top candidate over the runner-up. SCR ≈ 1 signals an ambiguous decision; large SCR signals a confident one.*

### 6.3 The bridge: conditional expectation

The two maps are joined per condition: the spectral state $s(\theta)$ is paired with the classifier's category proportions $q(\theta)$ on the shared key (seed, layer, noise, modification). The central object of the spectral analysis is the spectral signature of each error category — the conditional expectation of the spectral observables given that the classifier assigned that category,

$$\mu_c = \mathbb{E}[S(\mathrm{P}) \mid C = c] \in \mathbb{R}^6. \quad \text{(N6.9)}$$

Its empirical estimator restricts to the conditions whose dominant (most frequent) error is c, $\Theta_c = \{\theta : \arg\max_{c'} q_{c'}(\theta) = c\}$, and averages:

$$\hat{\mu}_c = \frac{1}{|\Theta_c|} \sum_{\theta \in \Theta_c} s(\theta). \quad \text{(N6.10)}$$

The spectral signature answers the interpretive question "what does the output distribution look like when the model commits error type c?" To compare error types, the analysis uses the standardized spectral distance — the Euclidean distance between signatures after z-scoring each observable across categories (so that the large-scale SCR axis does not dominate),

$$d(c, c') = \|z(\mu_c) - z(\mu_{c'})\| = \sqrt{\sum_{j=1}^{6} \left(z_j(\mu_c) - z_j(\mu_{c'})\right)^2}. \quad \text{(N6.11)}$$

A small distance means two categories arise from nearly the same distributional regime; a large distance means distinct regimes. This is how the spectral analysis interprets and differentiates the error types — it characterizes their nature and quantifies their mechanistic similarity, rather than classifying responses anew.

### 6.4 Why the relationship is one-directional

The signature map $c \mapsto \mu_c$ is always well defined: a conditional expectation of an integrable observable exists for every category. The reverse — recovering the category from the spectral state — is, in general, ill-posed. The argument is short.

Consider the Bayes assignment $\hat{c}(s) = \arg\max_c P(C = c \mid s)$ induced by any classifier that uses only $s$. Because a single distributional shape can be produced by responses of different categories — a sharply peaked distribution may yield a correct response or a confident semantic substitution with the same $(\mathrm{SF}, \mathrm{PR}, \mathrm{SCR})$ — the map from category to spectral support is many-to-one. Equivalently, the conditional law $P(C \mid s)$ is not concentrated for the spectrally overlapping categories.

Hence there exist categories c for which

$$P(C = c \mid s) < \max_{c' \neq c} P(C = c' \mid s) \quad \text{for all s}, \quad \text{(N6.12)}$$

so that $\hat{c} \equiv$ c is never selected: its recall, and therefore its F1, are identically zero. This is exactly the degeneracy observed when one trains a logistic regression $s \to C$ on the six observables: the spectrally extreme categories (correct, neologism, no-response) are recovered while the overlapping middle categories (semantic, mixed, formal, nonword, unrelated) collapse to zero. The failure is not numerical; it is the signature of forcing a one-directional relationship to run backward.

The two directions are therefore not symmetric:

$$C \to \mathbb{E}[S \mid C] \quad \text{well} - \text{posed (characterization)} \qquad S \to C \quad \text{ill} - \text{posed (pseudo} - \text{classification)}. \quad \text{(N6.13)}$$

The classifier supplies the labels; the spectral map supplies, for each label, the distribution it came from. Reversing the arrow manufactures a redundant and degenerate second classifier, which is why the paper retains exactly one classifier.

### 6.5 Mechanistic reading of the observables

Because the observables describe the generative state, the signatures admit a direct mechanistic reading. The table summarizes the correspondence used throughout the paper.

| **Observable (high value)** | **Distributional meaning** | **Error-type signature** |
|---|---|---|
| SF, H high | flat, uncertain distribution | neologism, no-response (lexical access collapsed) |
| α, SCR high | single sharp peak, confident | correct; moderate values → semantic competition |
| PR large | many effective competitors | graded semantic field; PR small → one token dominates |
| $D_{KL}$ large | committed away from uniform | decision converged to a few candidates |

*Supplementary Table S7. Mechanistic reading of the spectral observables. The classifier states which error occurred; the spectral signature states the distributional regime that produced it. Columns: the first names an observable and the direction of the reading, high meaning large relative to the range that observable takes across the perturbation grid rather than against any absolute threshold; the second describes the shape of the output distribution that produces such a value; the third names the error types that arise from that shape. The rows are readings of the same distribution from different angles and are not mutually exclusive.*

### 6.6 Consequence for the paper's structure

All spectral results in the paper are instances of the well-posed direction $C \rightarrow \mathbb{E}[S \mid C]$: the spectral hierarchy (Fig. 1) orders the signatures $\mu_c$ by flatness; the spectral distance matrix (Fig. 3) tabulates $d(c, c')$; and the co-occurrence analysis (Fig. 4) relates that distance to how often two categories appear together. None of these requires a classifier other than $C$; each is a characterization keyed on its labels. The spectral observables are, in the paper's words, not classifier labels but a quantitative characterization of the distribution from which the erroneous output was sampled.